\documentclass[acmsmall,nonacm]{acmart}
\AtBeginDocument{%
  \providecommand\BibTeX{{%
    \normalfont B\kern-0.5em{\scshape i\kern-0.25em b}\kern-0.8em\TeX}}}

\setcopyright{none}

\usepackage{url}
\usepackage{amsmath}

\usepackage{amssymb}
\usepackage{graphicx}
\usepackage{textcomp}
\usepackage{color, colortbl}
\usepackage{xcolor}
\usepackage{caption}
\usepackage{subfigure}
\usepackage{multirow}
\usepackage{algpseudocode}
\usepackage[vlined,ruled,linesnumbered]{algorithm2e}
\usepackage{xspace}
\usepackage{enumitem}
\usepackage{tcolorbox}
\usepackage{tikz}
\usepackage{pifont}
\usepackage{booktabs}

\DeclareMathOperator*{\argmax}{arg\,max}

\makeatletter
\DeclareRobustCommand\onedot{\futurelet\@let@token\@onedot}
\def\@onedot{\ifx\@let@token.\else.\null\fi\xspace}
\def\eg{\emph{e.g}\onedot} 
\def\ie{\emph{i.e}\onedot}

\def\etal{\emph{et al}\onedot}
\makeatother

\definecolor{dg}{rgb}{0,0.694,0.298}
\definecolor{purple}{rgb}{0.4,0.176,0.569}
\definecolor{iris}{rgb}{0.35, 0.31, 0.81}
\definecolor{tabgray}{rgb}{0.85,0.85,0.85}

\newcommand{\qing}[1]{\textcolor{black}{#1}}

\newcommand{\revise}[1]{{\color{black}#1}}

\newtheorem{definition}{Definition}
\def\BibTeX{{\rm B\kern-.05em{\sc i\kern-.025em b}\kern-.08em
    T\kern-.1667em\lower.7ex\hbox{E}\kern-.125emX}}

\begin{document}

\title{NPC: \underline{N}euron \underline{P}ath \underline{C}overage via Characterizing Decision Logic of Deep Neural Networks}

\author{Xiaofei Xie}
\email{xiaofei.xfxie@gmail.com}
\authornotemark[1]
\affiliation{%
  \institution{Singapore Management University}
  \country{Singapore}
}

\author{Tianlin Li}
\email{tianlin001@e.ntu.edu.sg}
\affiliation{%
  \institution{Nanyang Technological University}
  \country{Singapore}
}
\authornote{ Equal contribution.}

\author{Jian Wang}
\email{jian.wang@ntu.edu.sg}
\affiliation{%
  \institution{Nanyang Technological University}
  \country{Singapore}
}

\author{Lei Ma}
\email{malei@ait.kyushu-u.ac.jp}
\affiliation{%
  \institution{Kyushu University}
  \country{Japan}
}

\author{Qing Guo}
\email{qing.guo@ntu.edu.sg}
\authornotemark[2]
\affiliation{%
  \institution{Nanyang Technological University}
  \country{Singapore}
}

\author{Felix Juefei-Xu}
\email{juefei.xu@gmail.com}
\affiliation{%
  \institution{Alibaba Group}
  \country{USA}
}

\author{Yang Liu}
\email{yangliu@ntu.edu.sg}

\authornote{Corresponding Authors.}
\affiliation{%
  \institution{Nanyang Technological University, Singapore,}
  \institution{Zhejiang Sci-Tech University}
  \country{China}
}


\begin{abstract}
Deep learning has recently been widely applied to many applications across different domains, \eg, image classification and audio recognition. However, the quality of Deep Neural Networks (DNNs) still raises concerns in the practical operational environment, which calls for systematic testing, especially in safety-critical scenarios. Inspired by software testing, a number of structural coverage criteria are designed and proposed to measure the test adequacy of DNNs. However, due to the blackbox nature of DNN, the existing structural coverage criteria are difficult to interpret, making it hard to understand the underlying principles of these criteria. The relationship between the structural coverage and the decision logic of DNNs is unknown. Moreover, recent studies have further revealed the non-existence of correlation between the structural coverage and DNN defect detection, which further posts concerns on what a suitable DNN testing criterion should be. 
%

In this paper, we propose the \textit{interpretable} coverage criteria through constructing the decision structure of a DNN. Mirroring the control flow graph of the traditional program, we first extract a decision graph from a DNN based on its interpretation, \qing{where a path of the decision graph represents a decision logic of the DNN.} Based on the control flow and data flow of the decision graph, we propose two variants of path coverage to measure the adequacy of the test cases in exercising the decision logic. The higher the path coverage, the more diverse decision logic the DNN is expected to be explored. Our large-scale evaluation results demonstrate that: the path in the decision graph is effective in characterizing the decision of the DNN, and the proposed coverage criteria are also sensitive with errors including natural errors and adversarial examples, and strongly correlated with the output impartiality.

\end{abstract}


\maketitle
\section{Introduction}
\begin{figure*}
    \centering
    \includegraphics[width=0.8\linewidth]{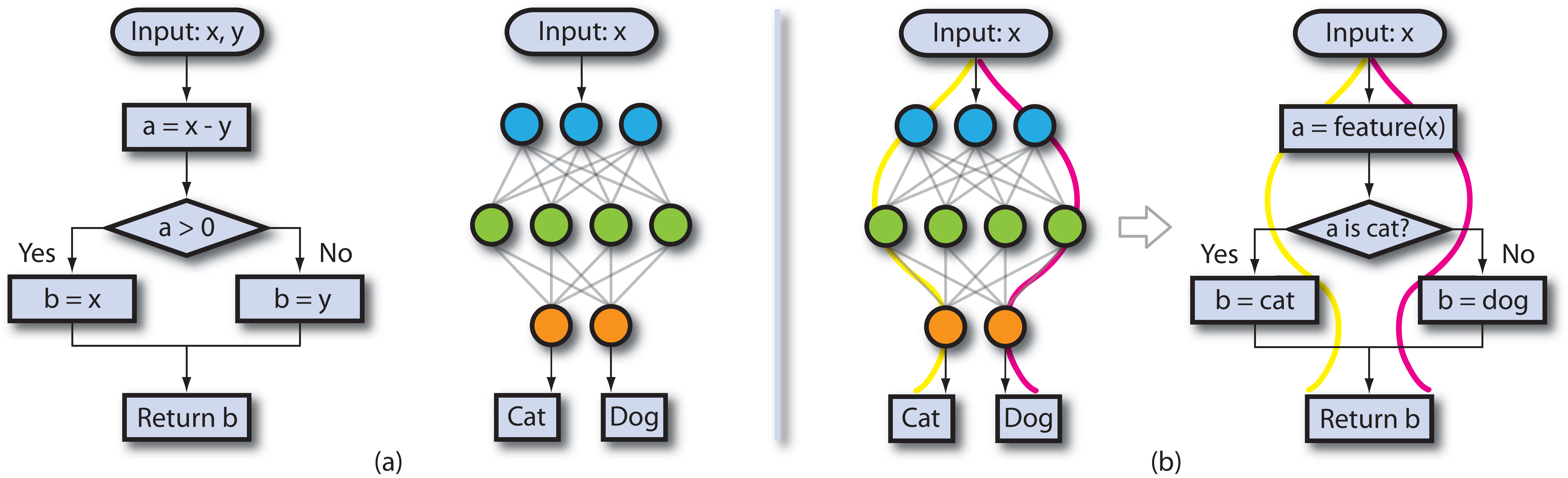}
    \caption{(a) Difference between traditional software and DL software, (b) Paths in the DL software.}
    \label{fig:motivating}
\end{figure*}
\qing{Deep Learning (DL) has achieved tremendous success and demonstrated its great potential in solving complex tasks in many cutting-edge real-world applications such as image classification~\cite{app_img}, speech recognition~\cite{app_speech}, natural language processing~\cite{app_nlp} and software engineering tasks (\eg, commit message generation~\cite{commit_gen}, vulnerability detection~\cite{NIPS2019_9358} and clone detection~\cite{deckard})}.
We have also seen an increasing demand to apply DL in some safety-critical areas, such as autonomous driving~\cite{ autodriving}, healthcare~\cite{medi_cancer} and security applications~\cite{lin2021phishpedia}. However, Deep Neural Networks (DNNs) have been demonstrated with quality issues, \eg, to be vulnerable to adversarial attacks in spite of having achieved high accuracy~\cite{biggio2013evasion}. \qing{We have already witnessed real-world cases caused by quality issues of DNNs}, \eg, Tesla/Uber accidents~\cite{uber_crash} and incorrect diagnosis in healthcare~\cite{medi_cancer}. \qing{The systematic assessment of the quality of DNNs is urgently needed, especially 
in safety- and security-critical scenarios.}

Even until recently, the common way of quality evaluation of DNNs still mostly relies on train-validation-test splits to calculate the accuracy, which may over-estimate the performance of the model~\cite{ribeiro-etal-2020-beyond}. \qing{This motivates the software engineering community to propose a variety of novel approaches and tools for systematically testing the quality of the DL software (\ie, the DNN).} However, due to the fundamental difference between the traditional code-based software and the DL software (see Fig.~\ref{fig:motivating}(a)), the software testing techniques (\eg, code coverage) could not be directly applied in DL software, making testing of the DL software be a new challenge. 

To this end, inspired by the white-box testing of the traditional software, a number of approaches towards testing of DL systems have been recently proposed~\cite{nerualmisled}. Specifically, some structural coverage criteria are proposed to measure the test adequacy of the test suite via analyzing the internal neuron activities, \eg, Neuron Coverage (NC)~\cite{pei2017deepxplore}, its variants (\eg, KMNC, NBC, SNAC)~\cite{ma2018deepgauge}, the combinatorial testing based criteria~\cite{ma2019deepct,sekhon2019towards},
the Surprise Adequacy~\cite{test:surprise} and etc.
Based on the coverage criteria, automated testing techniques (\eg, DeepXplore~\cite{pei2017deepxplore}, DeepTest~\cite{tian2018deeptest} and DeepHunter~\cite{test:deephunter}) are further developed to generate test cases for maximizing the coverage. 


\qing{DNN testing aims to cover diverse internal decision logics of DNN such that more bugs (\ie, incorrect decisions) could be detected~\cite{sekhon2019towards,zhang2020machine}. }
Due to the black-box nature and complexity of DNNs, \qing{it is unknown and hard to understand the underlying principles of existing coverage criteria~\cite{nerualmisled}}. Specifically, for traditional software, the defects reside on particular instances of the program’s control and data flow~\cite{ma2019test}. Thus, the code-based coverage criteria (\eg, statement coverage, branch coverage) are meaningful and essential as covering more statements or branches is more likely to capture diverse behavior (\ie, program logic) in terms of the control and data flow. \qing{However, unlike the traditional software, the decision logic of DNN is opaque to human, making it hard to understand the relationship between the decision logic and existing coverage criteria}. \qing{For example, it is not clear whether (de-)activating more neurons (\ie, NC) is better to capture more diverse behaviors of the DNN.} \qing{Thus, one may wonder what is the semantics and meaning behind existing coverage criteria.  }
  
Some recent research~\cite{nerualmisled,dong2019limited,ncmislead,fsecorrelation} started to perform a more in-depth investigation of existing coverage criteria and have demonstrated their limitations in different perspectives. 
Harel-Canada et al. have shown that NC is neither positively nor strongly correlated with defect detection, naturalness and output impartiality ~\cite{ncmislead}. For example, their experimental results have shown that only a few inputs could already achieve 100\% NC~\cite{sekhon2019towards}. Li et al. demonstrated that the existing testing criteria are not effective or sensitive in distinguishing same-size test sets of \textit{natural} examples including different levels of misclassification rates~\cite{nerualmisled}. Dong et al. further showed that there is a limited correlation between the coverage criteria and the robustness of the DNN~\cite{dong2019limited}.  
These results force us to rethink what could be a better coverage criterion of DNN and how could we design more interpretable coverage criteria.

Motivated by the aforementioned challenges and limitations, in this paper, we propose coverage criteria from the new angle, called Neuron Path Coverage (NPC). We argue that a better coverage criterion should be more relevant to the decision logic of the DNN. To this end, NPC is designed to be \textit{interpretable}\footnote{{We say that NPC is interpretable as it is designed to identify the key neurons that contribute more on the decision of the model based on the interpretation technique, \ie, Layer-Wise Relevance Propagation (LRP)~\cite{bach2015pixel}}} (\ie, easily understandable) and can better characterize the decision structure of the DNN.
\qing{In traditional software, a path in the control flow graph could represent the program logic of the software (\eg, login, logout, invalid user input checking and etc.).}
Inspired by this, we extract \textit{Critical Decision Paths} (CDPs) from the DNN, which represent the internal decision of the DNN on given inputs. For example, in Fig.~\ref{fig:motivating}(b), the DNN makes the decision along the corresponding CDPs (\eg, the yellow path or the red path), which is to classify the input as a dog or a cat. 
Based on the interpretation technique~\cite{bach2015pixel}, we extract \textit{critical} neurons in each layer, which dominate the decision of the DNN on the target input. The critical neurons between the layers form the CDP. Compared with the existing neuron-based coverage metric (\eg, NC~\cite{pei2017deepxplore}, $k$-multisection Neuron Coverage~\cite{ma2018deepgauge}, IDC~\cite{IDCcov}),
CDP considers not only the critical neurons in one layer but also the relationships among layers. Moreover, the CDP is clearly related to decision-making.
We then recover the learned decision logic of the DNN by extracting CDPs from all training samples. Intuitively, the CDP can be regarded as the control flow of the DNN that shows the important decision logic. Hence, the objective of DL testing is to generate diverse test inputs that can trigger different decision logic (\ie, trigger different control/data flow of the DNN). We firstly construct a Decision Graph (DG) to represent the overall decision logics based on the training data. Since the number of training samples may be very large, we propose the path abstraction technique to build the abstract CDP. The abstract state represents the common decision logic that is shared by some inputs. 
Based on the DG, we define two variants of path-based coverage criteria, named Structure-based Neuron Path Coverage (SNPC) and Activation-based Neuron Path Coverage (ANPC). Specifically, SNPC is designed based on the control-flow of the DNN (\ie, neurons in the path) while ANPC is designed based on the data-flow of the DNN (\ie, neuron activation values in the path).
An input can increase the coverage if it can trigger a different CDP or the similar CDP but with different activation values.

\qing{To demonstrate the effectiveness of our technique, we evaluate NPC on 4 widely-used datasets.} Specifically, we first evaluate the accuracy of the extracted paths, \ie, whether they are critical and useful in explaining the decision of the DNN. Our results show that the CDPs are critical for the prediction of inputs. \qing{After masking the outputs along the CDPs, the prediction results are changed greatly (\eg, 98.9\% inconsistency rate on average).} Moreover, the CDP is accurate to distinguish decision behaviors between different inputs in the same class. Then, we evaluate the usefulness of the proposed coverage criteria and the results show that NPC could distinguish the same-size test sets including a different number of natural errors. In addition, NPC is positively correlated with the output impartiality~\cite{ncmislead}, \ie, test set with high output bias has lower coverage. We also evaluate the efficiency of the coverage calculation, and the results show that the time overhead is not very expensive although NPC adopts the interpretation analysis. 

In summary, this paper makes the following main contributions:
\begin{itemize}
\item We propose a method to construct a decision structure of the DNN based on the CDP extraction and abstraction.
\item Based on the decision structure, we propose two variants of coverage criteria that are more relevant to the decision-making logic of the DNN. 
\item We conduct the comprehensive evaluation and demonstrate the effectiveness of the proposed approach. Our results show that the CDP is closely relevant to the decision-making; The proposed coverage criteria (\ie, SNPC and ANPC) are positively correlated with the detection of natural errors and output impartiality, respectively.
\end{itemize}


\section{Background and Motivation}\label{sec:motivation}
\qing{In this section, we briefly introduce software testing, the challenges of DL testing, and the interpretation technique LRP.}

\subsection{Traditional Software Testing}
Software testing has been widely studied in the past decades to evaluate the quality of the software. 
A central question of software testing is ``what is a test criterion?'', which defines what constitutes an adequate test~\cite{pathcov}.
In traditional software, the program logic is usually described by the programmer and can be represented as the control flow graph (see Fig.~\ref{fig:motivating}(a)). Based on the clear and easy-to-understand logic representation in the program, different coverage criteria have been proposed to measure the test adequacy, \eg, \textit{ statement coverage, branch coverage}, and \textit{path coverage}. These criteria are related to the program logic and the underlying principles are also intuitive: more program logic or behaviors could be explored by increasing the coverage (\eg, executing new statements or paths of the program).

\subsection{Challenges and Motivation}
Deep Neural Networks (DNNs) consist of multiple layers of neurons, which perform the nonlinear calculation for feature extraction and transformation. Similar to traditional software testing, the goal of DL testing is to test the decision logic of the DNN such that the incorrect prediction could be detected~\cite{sekhon2019towards}. However, unlike the control flow graph of the program, the logic of the DNN is opaque to humans, making it hard to design suitable coverage criteria. 
\qing{Neuron outputs of the DNN could reveal the behavior of the DNN, thus some neuron outputs based coverage criteria~\cite{pei2017deepxplore,ma2018deepgauge,test:surprise} are proposed.} However, it is not immediately clear how the neuron outputs reflect the internal decision logic, making it hard to understand its underlying principles and the meaningfulness of existing coverage criteria~\cite{nerualmisled}. 

The value range of neuron outputs is huge and it is often impractical to enumerate all possibilities. For efficiency, existing coverage criteria adopt different strategies to reduce the testing space. For example, NC~\cite{pei2017deepxplore} only considers two states (\ie, activated/inactivated) for the output of each neuron. $k$-multisection neuron coverage~\cite{ma2018deepgauge} discretizes the output value to $k$ buckets. Surprise metrics~\cite{test:surprise} select one of the layers to calculate the surprise score. The limitation is that it is unclear about the relationship between the coverage criteria and the decision logic of the DNN (\eg, whether a new decision-making behavior could be triggered by activating one neuron in NC). Consequently, if the coverage criteria are too coarse, some decision logic can be missed. For example, since NC is too coarse, several inputs can achieve very high neuron coverage~\cite{ncmislead,sekhon2019towards}. If the coverage criteria are too fine-grained, a huge number of inputs are needed. It is because many of them would trigger very similar decision logic with little difference.
The further results~\cite{sekhon2019towards,ncmislead,nerualmisled,dong2019limited} have also demonstrated the limitation of the existing coverage criteria, which call for new testing criteria design.
\qing{ 
Overall, this work is motivated by the recent findings ~\cite{ncmislead,nerualmisled,sekhon2019towards} on the limitation of the existing coverage criteria: \ding{172} the existing coverage criteria lack the interpretability and are unaware of the decision logic~\cite{nerualmisled,sekhon2019towards}, \ding{173} the existing coverage criteria are not sensitive to the defect detection including both of adversarial examples and natural errors~\cite{nerualmisled,ncmislead} and \ding{174} the existing coverage is not positively correlated with the output diversity~\cite{ncmislead}. 
}

\noindent \fbox{
	\parbox{0.95\linewidth}{\textbf{Motivation}: Similar to the traditional software testing, a better coverage criterion for DL testing should be \textit{interpretable} and more closely \textit{related} to the decision logic of the DNN, which is what the existing coverage criteria lack. When new inputs increase the coverage, it is expected that new decision logic would be exercised and covered.
	}
}




\subsection{Layer-Wise Relevance Propagation (LRP)}\label{sec:lrp}

\begin{figure*}
    \centering
    \includegraphics[width=.7\linewidth]{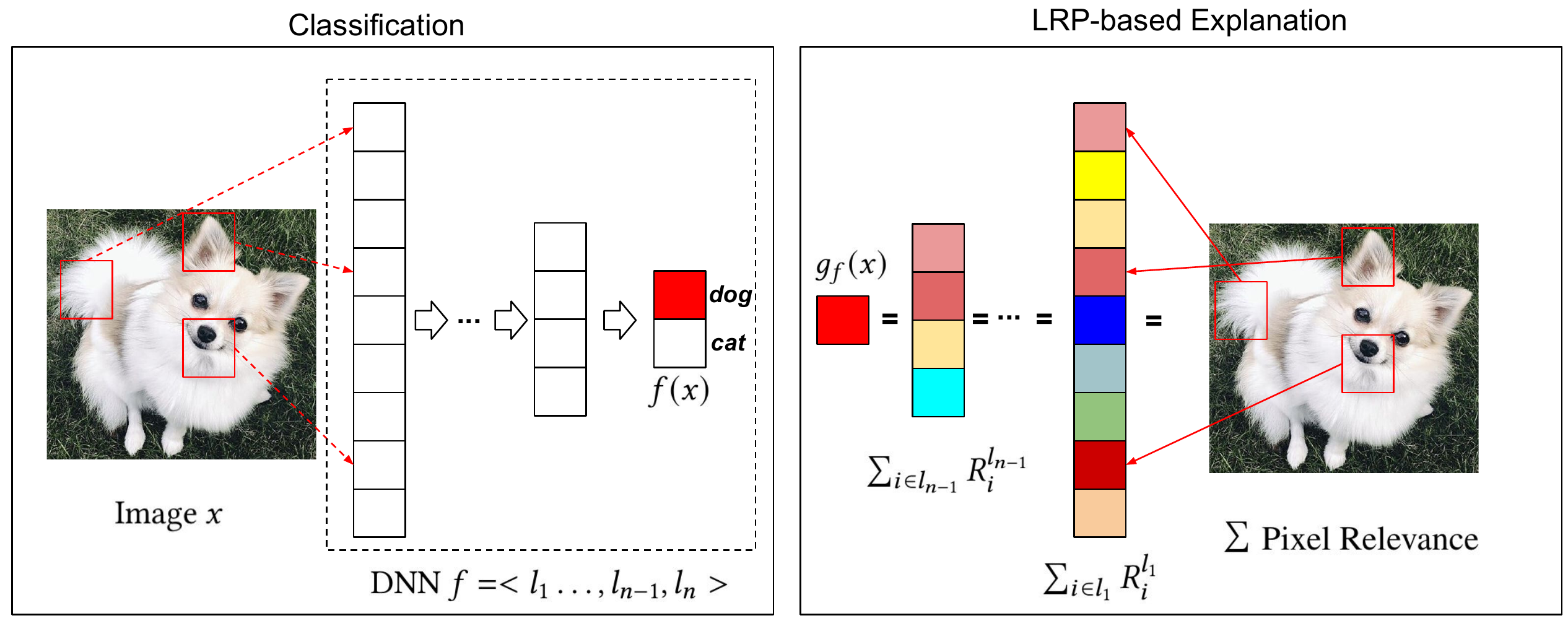}
    \caption{\revise{Visualization of the LRP Process \cite{bach2015pixel}}}
    \label{fig:lrpvis}
\end{figure*}

\revise{
\begin{definition}
A Deep Neural Network (DNN) $f$
consists of multiple layers $\left\langle l_1,\ldots,l_n\right\rangle$, where $l_1$ is the input layer, $l_n$ is the output layer, and $l_2,\ldots,l_{n-1}$ are hidden layers. The inputs of each layer are from outputs of the previous layer.
\end{definition}

In this work, we mainly focus on the classifier $f:X \rightarrow Y $, where $X$ is a set of inputs and $Y$ is a set of classes.  Given an input $x\in X$, $f(x)=\argmax_{y}O_n$, where $O_n$ is a $|Y|$-dimensional vector, which is the output of the output layer $l_n$. The input is classified as the class $y$, and we use $g_f(x)=O_n(y)$ to represent the output value (\eg, the logits) for the class $y$.
Due to the complex non-linear calculation in the DNN, it is usually hard to understand the decision.

Layer-Wise Relevance Propagation (LRP)~\cite{bach2015pixel} is an effective way to interpret the decision by calculating the key features in the input. Given a $d$-dimensional input $x$ and the classifier $f$, $x$ is classified as the class $y$. To interpret the decision, LRP calculates the relevance (or contribution) of each dimension (\ie, pixel) in the input. Specifically, the classifier $f$ outputs $g_f(x)$ on the class $y$ and LRP calculates the relevance of each input dimension:
$$g_f(x) = \sum_{i=1}^{d} R_i^{l_1}$$
where $R_i^{l_1}$ represents the relevance of the dimension $i$ in the input layer $l_1$ (\eg, the pixel in an image). The sum of the relevance scores is equal to the output value $g_f(x)$, which means that each dimension can contribute to the final decision of predicting $y$ (\ie, $g_f(x)$). Intuitively, $ R_i^{l_1}<0$ contributes evidence against the presence of the class which is to be classified while $ R_i^{l_1}>0$ contributes evidence for its presence~\cite{bach2015pixel}.

The relevance is calculated by a layer-by-layer backward propagation (from the output layer $l_n$ to the input layer $l_1$) such that:
$$g_f(x)=R_y^{l_n} = \sum_{i\in{l_{n-1}}} R_i^{l_{n-1}} = \ldots = \sum_{i\in{l_1}} R_i^{l_1}$$
where $R_i^{l}$ represents the relevance of the neuron $i$ at the layer $l$. The relevance calculation starts from the output neuron $y$ at the last layer $l_n$ (\ie, $R_y^{l_n}$) until the relevance $R_i^{l_1}$ of the input layer is calculated. At each layer, the sum of the relevance of all neurons is equal to the output $g_f(x)$. To be more specific, each relevance is calculated based on the following relation:  
$$R_j^{(l_{m+1})} = \sum_{i\in(l_m)} R_{i\leftarrow j}^{(l_m,l_{m+1})}, R_i^{(l_m)} = \sum_{j\in(l_{m+1})} R_{i\leftarrow j}^{(l_m,l_{m+1})}$$
where $R_j^{(l_{m+1})}$ represents the relevance of the neuron $j$ at the layer $l_{m+1}$, $R_{i\leftarrow j}^{(l_m,l_{m+1})}$ represents the relevance of the neuron $i$ at the layer $l_m$, where the output of the neuron $i$ is the input of the neuron $j$ at the layer $l_{m+1}$.
The relevance calculation starts from the output neuron at the last layer $l_n$ is $R_y^{(l_n)}=g_f(x)$ until the relevance $R_i^{(1)}$ of the first layer (\ie, the input layer) is calculated. 

Fig.~\ref{fig:lrpvis} shows the LRP calculation process on an image. In the left box, we show the ``black-box'' prediction where the image is classified as a dog and outputs $g_f(x)$. 
To interpret why it is classified as the dog, LRP calculates the relevance layer-by-layer starting from the output $g_f(x)$ (shown in the right box). Finally, neurons/pixels that have more impact on the prediction are identified. \qing{Note that different colors represent the different relevance scores of the corresponding neurons. Here we use the red colored squares to represent the neurons that are more critical for extracting features of dogs.}
}
\section{Methodology}\label{sec:methodology}
\begin{figure*}
    \centering
    \includegraphics[width=1\linewidth]{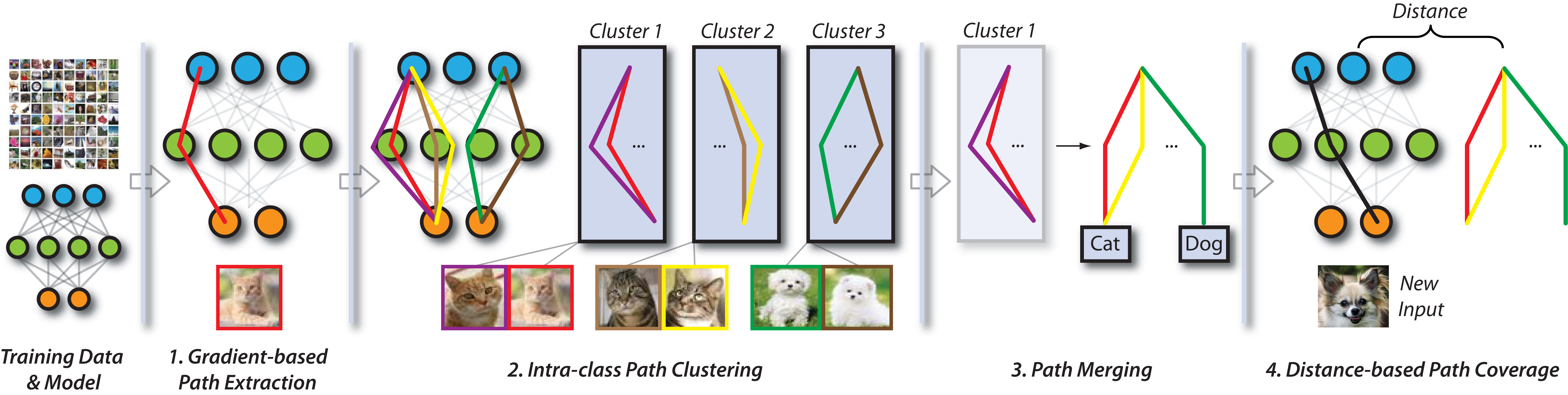}
    \caption{Overview of this work.}
    \label{fig:overview}
\end{figure*}

\subsection{Overview}
To design coverage criteria that are related to the decision logic of DNN, it is necessary to understand the decision-making of the black-box DNN (we say DNN is black-box since it is usually uninterpretable). \qing{Fortunately, some recent progress has been made on the interpretation of machine learning models~\cite{bach2015pixel,fong2017interpretable,wang2018interpret}, which can be used for the interpretable coverage criterion.}

In this work, we applied the \textit{Layer-Wise Relevance Propagation} (LRP)~\cite{bach2015pixel} to reveal the internal decision of an input.
LRP is an effective approach that interprets the classification decisions by the decomposition of nonlinear classifiers (\eg, DNNs) such that the key features in the input (\eg, pixels in the images) could be identified.
In particular, given an input, LRP calculates the \textit{relevance} of neurons in each layer starting from the last layer to the input layer in terms of the prediction result. The relevance represents the effect of neurons in the decision. \qing{Based on the relevance, we define the \textit{Critical Decision Path} (CDP) that is a set of critical neurons (\ie, with larger relevance in each layer) for the decision.} Consider the left figure in Fig.~\ref{fig:motivating}(b) as an example, the yellow path and the red path represent different decision logic for the prediction of dogs and cats, respectively. Our evaluation results show that similar inputs (\eg, dogs) tend to share a similar CDP (\ie, their decision logic is similar) while different inputs (\eg, dogs and cats) have different CDPs.

\qing{Inspired by the path coverage criterion in software testing, we propose the Neuron Path Coverage (NPC)}. For example, in the control flow graph of Fig.~\ref{fig:motivating}(a), it returns the maximum value of the inputs $x$ and $y$. The two paths show the clear logic of the program: the left path represents that $x$ is the larger one while the right path represents that $y$ is the larger one. Similarly, the extracted CDPs are used to show the logic of the DNN. For ease of understanding, we show a dummy graph in the right figure of Fig.~\ref{fig:motivating}(b), if the input executes the yellow path, it is classified as the cat, otherwise, a dog.

\qing{Fig.~\ref{fig:overview} shows a detailed workflow of our work.} The decision logic of the DNN is determined by the training data, given the target DNN, we first individually extract the CDP for each training data (Section~\ref{sec:pathextract}). 
The extracted CDPs represent the decision logic that is learned from the training data. Given an input, we measure the coverage by calculating the distance between the CDP of a given input and CDPs of training data (\ie, the difference of the decision logic). However, calculating the distance with all training samples is computationally expensive. To improve the efficiency, we perform the intra-class path clustering and abstraction on the CDPs of training samples (Section~\ref{sec:pathabstract}). 
Specifically, the assumption is that the samples with similar decision logic may have similar CDP. Thus, for the inputs predicted as the same class, we group their CDPs into $k$ clusters. It means that inputs with similar decision logic are clustered into one group. \qing{For example, similar cats are clustered together (\eg, the yellow cats and brown cats in Fig.~\ref{fig:overview}).}
For each cluster, we further propose the path abstraction that merges CDPs of inputs (in the cluster) into an abstract CDP. The abstract CDP represents the decision logic for all inputs in this cluster. Finally, the number of paths is greatly reduced. 
Based on the abstract CDPs, we propose the distance-based coverage criteria (Section~\ref{sec:criteria}), \qing{which measure to what extent a new decision logic is covered by the given test suite.} Specifically, we first identify the abstract CDP of the given input and then propose two coverage criteria. \qing{From the control flow perspective,} we calculate the \textit{similarity} between the new CDP and the abstract CDP of the corresponding cluster. From the data-flow perspective, we take the activation value of neurons in the abstract CDP into consideration and calculate the \textit{distance} (along the abstract CDP) between the input and training samples in the cluster. The similarity and distance reveal how different the decision behavior of the new input is.
Similar to~\cite{test:surprise}, we divided the similarity/distance to $m$ buckets and calculate how many buckets are covered by a given test suite. \revise{To adopt the coverage criteria in testing DNNs, we need to select the optimal hyperparameters for the CDP extraction and abstraction. We propose a method to select the parameters by evaluating the accuracy of the CDPs on affecting the decision. }




\subsection{Path Extraction}\label{sec:pathextract}
In this section, we will introduce how to extract the critical decision path for a given input.

\qing{As described in Section~\ref{sec:lrp}, the relevance $R_i^{l}$ represents the contribution to the final decision.} The larger relevance the relevance  $R_i^{l}$, the more contribution the 
neuron $i$ at the layer $l$. Since the DNN prediction is based on the features extracted and transformed layer-by-layer, we use the critical neurons with large relevance in each layer to represent the internal decision of the DNN. Then we define the critical decision path along all layers of the DNN as follows:

\begin{definition}[Critical Decision Path]\label{def:cdp}
Given an input $x$ and the DNN $f$ consisting of $n$ layers, we define its $\alpha$-Critical Decision Path ($\alpha$-CDP) as a sequence of neurons sets $p=\left\langle s_1,\ldots,s_{n-1}\right\rangle$, where $s_i$ is a set of critical neurons that have the largest relevance values at the layer $l_i$ and
$$(\forall j\in s_i. R_j^{l_i}>0) \wedge (\sum_{j\in {s_i}} R_j^{l_i}>\alpha \cdot g_f(x))$$
%
where $\alpha$ is a parameter that controls the number of the selected critical neurons at each layer. 
\end{definition}

The neuron is \textit{critical} if it provides a positive contribution to the decision (\ie, the positive relevance). We first sort neurons of each layer by the relevance values, and the CDP is composed of the most critical neurons at each layer. As a layer may have multiple critical neurons, we use the parameter $\alpha$ to prioritize the more critical ones. The smaller the $\alpha$, the fewer critical neurons are selected at each layer.
Other neurons that are not selected in the CDP are called \textit{Non-Critical Decision Path (NCDP)}.

We define the width of the CDP as the average ratio of neurons selected at each layer.
It is worth mentioning that, for different inputs, the width of the CDP may be different even with the same value for the parameter $\alpha$.
For each input, we select the CDP that has a large effect on the prediction while the NCDP (\ie, other neurons) has little impact. Furthermore, the width of the CDP (\ie, the selected neurons) should be as small as possible such that it can represent different decision logic accurately. Consider an extreme case like that, we could use all neurons of the network as the CDP, which does not lose any information. However, it cannot distinguish the decision logic between different inputs because all inputs have the same CDP (\ie, the DNN). Our evaluation demonstrated that CDPs with a suitable width could better reveal different predictions, \qing{\ie, inputs predicted to be the same class share similar CDPs while inputs predicted to be different classes have different CDPs.}

\subsection{Path Abstraction}\label{sec:pathabstract}
The decision logic of the DNN is determined by the training data, thus we recover its main decision logic that is learned from the training data. Given a classifier $f$ and its training data $T$, we construct the CDP for each sample in $T$, which are used to estimate the learned logic of the DNN roughly.
However, analyzing all CDPs of $T$ may introduce high computational costs especially when the size of $T$ is large. Moreover, some training data may share similar decision behaviors. To reduce the computational cost, we introduce a two-stage method to calculate the abstract CDPs.

\paragraph{Intra-Class Path Clustering} 
The assumption is that, the inputs that are predicted to be the same class, should have similar CDPs because the DNN has similar decisions on these inputs. Thus, our basic idea is to generate representative CDPs, which could characterize the common decision pattern on these inputs.

In particular, suppose the classifier $f$ performs a classification task with a set of classes $Y$, we group the training samples on the predicted class $y\in Y$, \ie, $T_y=\{x\in T | f(x)=y\}$. For each class $y$, although the decision logic on the inputs $T_y$ is similar, they may still have some slight difference. For example, the decision of the yellow cats may be slightly different from the brown cats. 
To capture more fine-grained decision logic in each class $y$, we cluster CDPs of the samples in $T_y$ into $k$ groups. 
We apply the $K$-Means clustering algorithm in this paper. Specifically, for each sample, we obtain a vector that represents the status of all neurons of the DNN. If a neuron is included in the CDP, its status is marked as 1 in the vector, otherwise, it is marked as 0. Note that all vectors have the same size such that they can be clustered.
$C_j^y\subseteq T_y$ denotes the training samples in the $j^{th}$ cluster. The samples in the same cluster have more similar decision-making logic than samples in other clusters.

\paragraph{Path Merging} 
For CDPs in each cluster, we then propose the path abstraction by merging them into an abstract path. Specifically, for $C_j^y$ in the $j^{th}$ cluster, we calculate an abstract CDP $\hat{p} = \left\langle \hat{s}_1,\ldots,\hat{s}_{{n-1}}\right\rangle$ by merging neurons in each layer:
$\forall i\in\{1,\ldots,n-1\}: \hat{s}_i = \bigcup_{x\in C_j^y} s_i^{x}$.
For each neuron $t\in \hat{s_i}$, we calculate its weight as:
$$w_t= \frac{|\{x | x\in C_j^y \wedge t \in s_i^{x}\}|}{|C_j^y|}$$
where $s_i^{x}$ is the set of critical neurons at the layer $i$ in the CDP of the input $x$.

Intuitively, for a set of inputs, the abstract CDP is the union of their critical neurons at each layer. The weight of each neuron $w_t$ represents the ratio of CDPs that the neuron $t$ belongs to. 
The larger the weight, the more critical this neuron is (\ie, it is critical for more CDPs of training samples in this cluster).
\qing{After merging the paths, the abstract CDP could be much larger than the CDP of one input.} There are some neurons in the abstract CDP, which may have small weight. Statistically, neurons with smaller weights may be less important for interpreting the decision on the set of inputs. Thus, we further introduce the parameter $\beta$ to select the critical neurons whose weights are above the threshold and obtain the abstract CDP $\hat{p}_\beta = \left\langle \hat{s}_{{1,\beta}},\ldots,\hat{s}_{{{n-1},{\beta}}}\right\rangle$, where
$$\hat{s}_{{i,\beta}} = \{t| t\in\hat{s}_i \wedge w_t >\beta \}$$

Based on the abstract CDPs, we characterize the decision logic of a DNN as a decision graph, which is similar to the control flow of programs.

\begin{definition}[Decision Graph]
The Decision Graph (DG) of a classifier $f$ is composed of a set of abstract CDPs $G= \{(\hat{p}_{\beta}^{1,1},\ldots,\hat{p}_{\beta}^{1,k}), \ldots, (\hat{p}_{\beta}^{n,1},\ldots,\hat{p}_{\beta}^{n,k})\}$, where $n$ is the number of the classes, $k$ is the number of the clusters for each class and $\hat{p}_{\beta}^{i,j}$ represents the corresponding abstract CDP with regards to the $j^{th}$ cluster of the class $i$.
\end{definition}
\subsection{Neuron Path Coverage}
\label{sec:criteria}
The decision graph represents the main logic of the DNN learned from its training data. 
Given a new input, it is expected that the input could be predicted correctly via the learned decision logic. 
DL testing aims to identify failed inputs that are possibly caused by triggering either \textit{unknown} logic or \textit{incorrect} logic. Specifically, \textit{unknown} logic means that the CDP of the input has low similarity with abstract CDPs in the DG, \ie, the DNN may not be able to learn the logic with the training data. \textit{Incorrect} logic means that the input exercises the incorrect CDP, \eg, one dog may have a similar CDP with a cat. We propose the path coverage to measure the adequacy of a given test suite. Intuitively, the more paths covered, the more logic explored.

Specifically, inspired by the traditional software testing, we consider the control flow and data flow of the DNN and propose two variants of path-based coverage criteria: the \textit{Structure-based Neuron Path Coverage (SNPC)} and the \textit{Activation-based Neuron Path Coverage (ANPC)}. \textit{SNPC} mainly considers the control flow, \ie, the path structure while \textit{ANPC} considers the data flow, \ie, the neuron outputs in the path. 

Theoretically, like paths in traditional programs, the number of the paths in DNN could be infinite. To mitigate this problem, we follow the strategy~\cite{test:surprise} by calculating the distance and use bucketing to discretize the space of the distance. 

\subsubsection{Structure-based Neuron Path Coverage}
Given an input $x$ with the CDP $p_x$, we first identify the clusters based on the predicted result $f(x)$. Suppose $G_{f(x)} =\{\hat{p}_{\beta}^{f(x),1},\ldots,\hat{p}_{\beta}^{f(x),k}\}$ is the set of abstract paths that are extracted from clusters of the class $f(x)$. We measure the similarity between $p_x$ and each of the abstract paths in $G_{f(x)}$, which represents the difference of the decision behavior of the given input $x$ with the existing abstract paths. Then we divide the similarity into multiple buckets and calculate the coverage, \ie, how many buckets are covered.

Given two paths $p_x$ and $\hat{p}$, we calculate the similarity layer by layer. For a layer $l$, the similarity is:
\begin{equation}\label{sec:jaccard}
    J_{p_x^l,\hat{p}^l}= \frac{s^x_l \cap \hat{s}_l}{s^x_l \cup \hat{s}_l}
\end{equation}
where $s^x_l$ and $\hat{s}_l$ are the neurons in the $l^{th}$ layer of $p_x$ and  $\hat{p}$, respectively.


Then we divide the continuous space of the similarity (\ie, [0,1]) into $m$ equal buckets $B=\{b_1,\ldots,b_m\}$. For an input $x$ and a target abstract path $\hat{p}$, the covered bucket at layer $l$ is denoted as:
$$b_{x, \hat{p}}^l = b_i \text{ if } J_{p_x^l,\hat{p}^l}\in (\frac{i-1}{m}, \frac{i}{m}]$$


Given a test suite $X$, we define the Structure-based Neuron Path Coverage as follows:
\begin{equation}
    SNPC(X)=\frac{|\{b_{x,\hat{p}}^l | \forall x\in X, \forall \hat{p} \in G_{f(x)}, \forall l \in f\}|}{n\cdot k \cdot |\hat{p}| \cdot m}
\end{equation}
 where $\hat{p}$ is the corresponding abstract CDP, $|\hat{p}|$ is the total number of layers, 
 $n$ is the number of the total classes, $k$ is the number of clusters in the class $f(x)$ and $m$ is the number of buckets. 

\revise{
Specifically, for each layer of the abstract path, we divide the neuron similarity [0, 1] (\ie, Jaccard similarity) between two layers into $m$ buckets. Finally, there are a total of $n \cdot k \cdot |\hat{p}| \cdot m$ buckets, where $n\cdot k $ is the total number of abstract CDPs and $|\hat{p}| $ is the number of layers. SNPC calculates how many buckets can be covered by a test suite $X$.
For a test case $x\in X$ and a layer $l$, we calculate the layer similarity between the CDP of $x$ and the corresponding abstract CDP $\hat{p}$. The similarity falls into a specific bucket $b_{x, \hat{p}}^l$. We identify all buckets covered by the test suite $X$, \ie, $\{b_{x,\hat{p}}^l | \forall x\in X, \forall \hat{p} \in G_{f(x)}, \forall l \in f\}$, and calculate the coverage.
}

\qing{We take the similarities with all $\hat{p}$ into consideration for a more fine-grained coverage analysis. This is because one test sample could be relevant with multiple abstract paths in the same class. 
For example, one abstract path may represent dogs with white color while another abstract path may represent dogs with long ears. Thus, comparing the similarity between the CDP of a test input and different abstract CDPs could measure the input on different feature levels (\ie, with different abstract CDPs).}

The higher the coverage criteria, the more diverse logic the DNN has been explored. However, \textit{SNPC} only considers the control flow, which might be coarse in some cases. To be more fine-grained, we therefore propose another variant of path coverage that adopts the concrete activation values of neurons in the path (\ie, data flow).

\subsubsection{Activation-based Neuron Path Coverage} 
In SNPC, we adopt the Jaccard similarity to calculate the structural similarity between the CDP of the input $x$ and the abstract CDP of a cluster. To be more fine-grained, we consider the data flow of the CDP, \ie, the activation values of neurons in the CDP and calculate the distance between the values of two CDPs. Specifically, given an input $x$ and a cluster $j$, we first select the closest training sample $x'$ from the cluster $j$, which has highest similarity with $x$: $$x' = \argmax_{x'\in C^{f(x)}_j} J_{p_x, p_{x'}} $$ where $J_{p_x, p_{x'}}$ is the Jaccard similarity between CDPs of $x$ and $x'$.
Then, we calculate the distance between the activation values of $x$ and $x'$ along the abstract CDP in the cluster. Intuitively, the larger the distance, the more different decision behaviors are triggered by the new input $x$, \ie, compared with the samples in cluster $j$, \qing{more behavior is covered by the new input.} For the input $x$, we use $A(x, \hat{p}^l)$ to denote the activation values of neurons in the $l^{th}$ layer of the abstract path $\hat{p}$. The distance at the $l^{th}$ layer between $x$ and $x'$ is defined as:
$$D^l_{x,x'}=||A(x, \hat{p}^l) - A(x', \hat{p}^l)||$$

\qing{Similar to~\cite{test:surprise}, we introduce an upper bound $U$ and divide the distance $[0,U]$ into $m$ buckets: $B=\{b_1,\ldots,b_m\}$}. Given an input $x$ and an abstract path $\hat{p}$ in the corresponding cluster, we define  $d_{x, \hat{p}}^l$ as the covered bucket, \ie, the layer distance $D^l_{x,x'}$ falls into a bucket: $$d_{x, \hat{p}}^l = b_i \text{ if } D^l_{x,x'} \in (U \cdot  \frac{i-1}{m}, U \cdot  \frac{i}{m}]$$

Given a test suite $X$, we define the Activation-based Neuron Path Coverage (ANPC) as:
\begin{equation} ANPC(X)=\frac{|\{d_{x,\hat{p}}^l | \forall x\in X, \forall \hat{p} \in G_{f(x)}, \forall l \in f\}|}{n\cdot k\cdot |\hat{p}|\cdot m}
\end{equation}

From the definitions, we can see that the key difference between SNPC and ANPC is that SNPC calculates the distance (\ie, $b_{x,\hat{p}_j}^l$) based on the path structure similarity while ANPC calculates the distance (\ie, $d_{x,\hat{p}}^l$) based on the neuron activation distance. With the coverage criteria, we could measure the adequacy of the test suite with regards to the decision behaviors.
\revise{
\subsubsection{Hyperparameter Tuning}
In the two coverage criteria, we have three hyperparameters, \ie, $\alpha$, $\beta$ and $k$, which determine the precision of the (abstract) CDPs  with regards to representing the decision logic. To adopt the proposed coverage criteria in DL testing, the hyperparameter tuning is required. We propose a strategy to evaluate the precision of the CDPs (under a set of hyperparameters) on representing the decision logic as follows:

\begin{enumerate}
    \item For each training data, we extract its CDP and the NCDP. The expectation is that the CDP has large impact on the decision of the data while the NCDP has less impact on the decision. Moreover, the width of the CDP \qing{should be} as small as possible such that only critical neurons are selected. Specifically, we respectively dropout the outputs of neurons (\ie, mask the results as zero) in the CDP and the NCDP, and evaluate whether the prediction result is changed (\ie, inconsistent). We evaluate the inconsistency rate on all training data on some hyperparameters and select the best ones that satisfies our assumption.
    \item Similarly, we evaluate the precision of abstract CDPs. For the abstract CDP of a cluster, we calculate the inconsistency rate on all training data of the cluster. For each data, we dropout the same neurons that belong to the abstract CDP. The inconsistency rate is expected to be high by masking the neurons of the abstract CDP and low by masking the neurons of the abstract NCDP.
\end{enumerate}
\qing{In our paper, the value ranges of the three hyperparameters $\alpha$, $\beta$ and $k$ are $\{0.7, 0.8, 0.9, 1.0\}$, $\{0.6, 0.7, 0.8, 0.9\}$ and $\{1, 4, 7\}$, respectively.}
}

\section{Evaluation}\label{sec:eval}

\qing{
\subsection{Setup}
To evaluate the effectiveness of our approach, we have implemented the coverage criteria based on the PyTorch framework. 
We design experiments to demonstrate the usefulness of our method in mitigating these challenges. Specifically, we aim to investigate and answer the following research questions:
}

\begin{itemize}[leftmargin=*]
     \item \textbf{RQ1}: How does the CDP affect the decision of the DNN? The CDP only selects the critical neurons in each layer. We design the experiment to evaluate whether such neurons are critical for DNN decisions.
    \item \textbf{RQ2}: How accurate is the abstract CDP in revealing internal decisions? Since the abstract CDP is extracted from CDPs of training samples in the corresponding cluster, we aim to evaluate whether the abstract CDP is still accurate in representing the DNN decision on these training samples in the cluster.
    \item \textbf{RQ3}: \qing{Is NPC sensitive with errors including natural errors and adversarial examples?} We investigate whether the proposed coverage criteria are sensitive in distinguishing the decision difference between erroneous samples and benign samples.
     \item \textbf{RQ4}: Is NPC correlated to the output impartiality? A test suite should exercise diverse behaviors and should not prefer only a few output values~\cite{nerualmisled}. We investigate whether the coverage criteria can reveal the impartiality of the test suite, \ie, the test suite with diverse behaviors should have high coverage while \qing{the test suite with biased behavior should have low coverage.}
    \item \textbf{RQ5}: How efficient is the calculation of NPC? Since NPC extracts decision behaviors based on the interpretation method, we investigate whether the interpretation analysis could introduce much time overhead for the coverage calculation. 
\end{itemize}

\subsection{Subject Datasets and DNNs}
We selected three widely used datasets in the image classification domain and trained four models with competitive test accuracy, which are widely used in the previous works~\cite{ma2018deepgauge,test:surprise,pei2017deepxplore}.

\begin{itemize}[leftmargin=*]
    \item \textbf{MNIST}~\cite{mnist} is for handwritten digit image recognition (\ie, handwritten digits from $0$ to $9$), containing $60,000$ training data and $10,000$ test data. 
    We select the five layer Convolutional Neural Network~\cite{test:surprise} (named \emph{SADL-1}) and achieve an accuracy of 99.1\%.
    \item \textbf{CIFAR-10}~\cite{cifar} is a collection of images for general-purpose image classification, including $50,000$ training data and $10,000$ test data in $10$ different classes~(\eg, airplanes, cars, birds, and cats). 
    We select two models, \ie, the 12-layer Convolutional Neural Network~\cite{test:surprise} (named \emph{SADL-2}) and VGG16, which achieved the accuracy of 90.36\% and 89.17\%, respectively.
    \item \textbf{SVHN}~\cite{netzer2011reading} dataset is house numbers in Google Street View images, including 73,257 training data and 26,032 test data. We select the model AlexNet that has an accuracy of 94.31\%.
    \item \revise{\textbf{ImageNet} is a large-scale visual recognition challenge~(ILSVRC) dataset. The ImageNet contains a large number of training data~(\ie, over one million) and test data~(\ie, 50,000) with 1000 classes, each of which is of size $224 \times 224 \times 3$. To reduce the complexity of training and CDP evaluation, we select the first 10 classes to train a classifier with the model VGG16 that performs the classification on the 10 categories. The model is trained with an accuracy of 76.81\%.} 

\end{itemize}
\subsection{Baselines}
We select the state-of-the-art coverage criteria as the baselines for the evaluation, \ie, Neuron Coverage (NC)~\cite{pei2017deepxplore}, $k$-Multisection Neuron Coverage (KMNC)~\cite{ma2018deepgauge}, Neuron Boundary Coverage (NBC)~\cite{ma2018deepgauge}, Likelihood-based Surprise Coverage (LSC)~\cite{ma2018deepgauge}, Distance-based Surprise Coverage (DSC)~\cite{test:surprise} and  Importance-Driven Coverage (IDC)~\cite{IDCcov}. Due to that the original IDC only supports Keras models, we implemented a PyTorch version of IDC for the comparison.

To be comprehensive, for NC, we set the threshold as 0, 0.2, 0.5 and 0.75 following~\cite{ncmislead}. For KMNC and NBC, we follow the configuration~\cite{ma2018deepgauge,test:deephunter} and set the $k$ value as 1,000 and 10. For LSC and DSC, we follow the configuration~\cite{test:surprise}, \qing{set the upper bound 2,000 and 2.0 respectively,} and set the number of buckets as 1,000. For the layer selection, we select the same layer in SADL-1 and SADL-2, which is used in~\cite{test:surprise}. For VGG16 in CIFAR and ImageNet, we select the last layer since it achieved the best result. For IDC, we follow the same configuration~\cite{IDCcov} and select the penultimate layer for each model. In the selected layer, we choose the top 12 important neurons. For SNPC and ANPC, we set the number of buckets as 200 and the upper bound as 2.0. 

\subsection{RQ1: Relationship between CDP and the DNN Decision}

\begin{table*}[!t]
    \centering

    \scriptsize
    \caption{The average width (\%) and inconsistency rate (\%) after masking neurons in the CDP and NCDP}


\begin{tabular}{cc|ccc|ccc|ccc|ccc}
\toprule 
 \multirow{2}{*}{Dataset} & \multirow{2}{*}{Model} & \multicolumn{3}{c|}{$\alpha=0.7$} & \multicolumn{3}{c|}{$\alpha=0.8$} & \multicolumn{3}{c|}{$\alpha=0.9$} & \multicolumn{3}{c}{$\alpha=1$}\tabularnewline
 &  & Width & Inc.C & Inc.NC & Width & Inc.C & Inc.NC & Width & Inc.C & Inc.NC & Width & Inc.C & Inc.NC\tabularnewline
\midrule 
\midrule 
MNIST & SADL-1 &  9.3 & 89.3 & 9.6 & \textbf{12.8} & \textbf{95.8} & \textbf{1.9} & 17.2 & 97& 0 & 33.3 & 98.9 & 1.8\tabularnewline
\midrule
\multirow{2}{*}{CIFAR} & SADL-2 & \textbf{20.9} & \textbf{100} & \textbf{0} & 27.2 & 100 & 0 & 36.3 & 100 & 0 & 63.4 & 100 & 0\tabularnewline
 & VGG16 & 10.2 & 100 & 12.1 & {13.3} & {100} & {1.5} & \textbf{17.9}& \textbf{100} & \textbf{0.1} & 33.8 & 100 & 0\tabularnewline
\midrule
SVHN & AlexNet & \textbf{23.0} & \textbf{100} & \textbf{2.1} & 29.6 & 100 &  1.2 & 38.2 & 100 & 0.2 & 64.6 & 100 & 0\tabularnewline
\midrule
\color{black}ImageNet &  \color{black}VGG16 &  \color{black}\textbf{41.1} &  \color{black}\textbf{99.3} &  \color{black}\textbf{12.5} &  \color{black}51.9 &  \color{black}99.8 &  \color{black} 13.9 &  \color{black}65.9 & \color{black}99.9 & \color{black}13.2 & \color{black}98.8 & \color{black}99.8 &\color{black}0.2
\tabularnewline
\bottomrule 
\end{tabular}

    \label{tab:singleinc}
\end{table*}
\subsubsection{Results of masking neurons of CDPs/NCDPs}
\paragraph{Mask all neurons.}
To evaluate whether the CDP has a critical impact on the decision of the DNN, \qing{we dropout the outputs of neurons in the CDP (\ie, mask them as zero)} and check whether the predicted result is inconsistent with the original result. Intuitively, we say that the CDP is accurate to reveal the decision if 1) the predicted output is largely affected after the neurons of the CDP are masked and 2) the predicted output is less affected after neurons in NCDP are masked. We extract CDPs for all training data and calculate the inconsistency rate after masking neurons in their CDPs and the corresponding NCDPs, respectively. 
\revise{
Specifically, the inconsistency rate of the classifier is defined as:
$$\frac{\{x |x\in X \wedge f(x)\neq f_m(x)\}}{|X|}$$
where $X$ is a set of test cases, $f(x)$ is the prediction result of the original classifier $f$ on the input $x$ and $f_m(x)$ is the prediction result after masking the CDP or NCDP of $x$. Intuitively, the higher the inconsistency rate, the larger the impact of the masked path.
}

Table~\ref{tab:singleinc} shows the average inconsistency rate by masking all neurons in the CDPs/NCDPs of all training data. The inconsistency rate shows the percentage of samples, whose prediction results are changed after the masking.  
To extract the path, we configure the parameter $\alpha$ (see Definition~\ref{def:cdp}) with different values, \ie, 0.7, 0.8, 0.9 and 1. Under each configuration, we show the average width of the CDP (Column \textit{Width}), which is the average ratio of neurons at each layer of the CDP, and the inconsistency rate after masking CDP (Column \textit{Inc.C}) and masking NCDP (Column \textit{Inc.NC}).

 The results show that the width of the CDP increases as the parameter $\alpha$ increases. For example, when $\alpha$ is 0.8, the average width on VGG-16 is 13.3\%, indicating that 13.3\% critical neurons are selected at each layer. When $\alpha$ is 1, the width becomes much larger (\ie, 33.8\%).
Furthermore, we found that almost all predicted results are changed when the CDPs are masked (\eg, for SADL-2 and VGG16, the inconsistency rate is 100\% for all configuration),
which demonstrates that the CDP is very \textit{critical} for the decision. Conversely, when we mask other neurons that are not in the CDP (\ie, the NCDP), the inconsistency is very low (0\% to 3\%). In particular, when $\alpha$ is small, the width is small (\ie, less neurons are selected) and the performance is relatively low. For example, when $\alpha$ is 0.7, the inconsistency rate of masking CDP on SADL-1 is 89.3\%, which is lower than 98.9\% ($\alpha=1$);
The inconsistency rate of masking NCDP in CIFAR-VGG16 is much higher (12.1\%) than other configurations regarding $\alpha$, which shows that there are some critical neurons (in NCDP) that are missed. 
Based on the results in Table~\ref{tab:singleinc}, in the following experiments, we set $\alpha$ as 0.8, 0.7, 0.9, 0.7 and 0.7 for MNIST-SADL-1, CIFAR-SADL-2, CIFAR-VGG16 and SVHN-AlexNet and Imagenet-VGG16, respectively. We also evaluate the CDPs on test data and the results show the similar trends.

\revise{
\paragraph{Mask parts of neurons of CDPs.} To further evaluate the impact of CDP, we conduct a fine-grained evaluation that only masks a small part of neurons based on their relevance. Specifically, for a given CDP, we sort the neurons at each layer of the CDP and then split these sorted neurons into 5 equal buckets that have different relevance values, \ie, the top 20\% neurons with high relevance, the 20\%-40\% neurons, the 40\%-80\% neurons and the last 20\% neurons (\ie, 80\%-100\%) with low relevance. We get 5 sub-CDPs that contain neurons with different relevance values. Then we mask each of the sub-CDPs and calculate the inconsistency rate.
Table~\ref{tab:maskpart} shows the inconsistency rate by only masking parts of the CDP. \qing{Overall, we could find that the neurons with higher relevance can have a larger impact on the prediction.} After we mask the most important part (\ie, the top 20\% neurons), the largest inconsistency rate is caused, which indicates that these neurons have larger impact. When the last 20\% neurons are masked, the inconsistency rate is the smallest.

We also observe that the results depend on the tasks and models. For example, on CIFAR-VGG16, after masking the top 20\% neurons of CDPs, most inputs are inconsistent with the original prediction (97.94\%), which shows that these neurons are quite important for the prediction. However, for the model SADL-2, only 9.13\% inputs are inconsistent after masking the top 20\% neurons of the CDPs. Refer to the results in Table~\ref{tab:singleinc}, after masking all neurons in the CDP ($\alpha=0.7$), the inconsistency rate on SADL-2 becomes 100\%. It means that all neurons of the CDPs on SADL-2 are important for the prediction. 
Consider the results of CIFAR-VGG16 and ImageNet-VGG16, although the same model architecture is used, their results are not consistent, indicating that the CDPs are different on different datasets.
}


\begin{table}[]
    \centering
   \small
    \caption{The average inconsistency rate (\%) after masking parts of neurons in the CDP.}
  {  \begin{tabular}{cc|ccccc}
\toprule
\multirow{2}{*}{Dataset} & \multirow{2}{*}{Model} & \multicolumn{5}{c}{Inconsistency Rate}\tabularnewline
\cline{3-7} \cline{4-7} \cline{5-7} \cline{6-7} \cline{7-7} 
 &  & 20\% & 20\%-40\% & 40\%-60\% & 60\%-80\% & 80\%-100\%\tabularnewline
 \midrule
 \midrule
MNIST & SADL-1 & 99.13 & 97.53 & 92.46 & 79.73 & 49.02\tabularnewline
 \midrule
\multirow{2}{*}{CIFAR} & SADL-2 & 9.13 & 7.65 & 3.29 & 2.39 & 0.58\tabularnewline
 & VGG16 & 97.94 & 84.13 & 62.42 & 43.21 & 22.64\tabularnewline
SVHN & AlexNet & 99.96 & 99.45 & 78.7 & 42.86 & 15.95\tabularnewline
 \midrule
ImageNet & VGG16 & 73.22 & 27.02 & 15.07 & 9.91 & 6.21\tabularnewline
\bottomrule
\end{tabular}
}
    \label{tab:maskpart}
\end{table}

\begin{table}
\centering
\scriptsize
\setlength\tabcolsep{4pt}
\caption{The average similarity of CDPs within a class or between different classes}

\begin{tabular}{cccccccccccccccc}
\toprule  
\multirow{2}{*}{$\alpha$} & \multicolumn{3}{c}{SADL-1} & \multicolumn{3}{c}{SADL-2} & \multicolumn{3}{c}{VGG16(CIFAR-10)} & \multicolumn{3}{c}{AlexNet} & \multicolumn{3}{c}{\color{black}VGG16(ImageNet)}\tabularnewline
\cline{2-16} \cline{3-16} \cline{4-16} \cline{5-16} \cline{6-16} \cline{7-16} \cline{8-16} \cline{9-16} \cline{10-16} \cline{11-16} \cline{12-16} \cline{13-16} 
 & Width & Intra & Inter & Width & Intra & Inter & Width & Intra & Inter & Width & Intra & Inter & \color{black}Width & \color{black}Intra & \color{black}Inter\tabularnewline
\midrule
\midrule
0.7 & 9.30\% & 57.7\% & 34.7\% & 20.9\% & 48.6\% & 23.2\% & 10.2\% & 86.5\% & 22.2\% & 23.0\% & 47.8\% & 30.0\%& \color{black}41.1\%& \color{black}54.9\%& \color{black}32.2\%\tabularnewline
0.8 & 12.80\% & 62.8\% & 39.1\% & 27.2\% & 56.5\% & 41.8\% & 13.3\% & 60.9\% & 26.0\% & 29.6\% & 54.5\% & 36.5\% & \color{black}51.9\% &\color{black} 60.8\% & \color{black}41.6\%\tabularnewline
0.9 & 17.20\% & 67.6\% & 45.3\% & 36.3\% & 64.4\% & 38.0\% & 17.9\% & 67.1\% & 32.0\% & 38.2\% & 63.2\% & 46.1\% & \color{black}65.9\% & \color{black}70.9\% & \color{black}55.8\%\tabularnewline
1 & 33.30\% & 74.5\% & 56.3\% & 63.4\% & 86.5\% & 62.0\% & 33.8\% & 78.3\% & 56.1\% & 64.6\% & 78.3\% & 75.4\%& \color{black}98.8\%& \color{black}99.9\%& \color{black}97.9\%\tabularnewline
\bottomrule 
\end{tabular}

\label{tab:distance}
\end{table}

\subsubsection{Results on effect of the width.}

The width of the CDP may affect its accuracy
in representing the decision logic. If the width is too large, the CDP contains some non-critical neurons that may also be included in other CDP, making it hard to distinguish them. \revise{We investigate the similarity between different classes. Specifically, for each class, we randomly select 100 test data that is correctly predicted as the target class. Then we calculate the average similarity of CDPs in the same class or between different classes as follows:
$$Intra=avg(\{sim(x^y_i, x^y_j) | y\in Y \wedge i\neq j \wedge x^y_i\in X^y \wedge x^y_j\in X^y\}) $$
$$Inter=avg(\{sim(x^{y_m}_i, x^{y_n}_j) | y_m\in Y \wedge y_n\in Y \wedge m\neq n \wedge x^{y_m}_i\in X^{y_m} \wedge x^{y_n}_j\in X^{y_n}\}) $$
where $Y$ is the set of all classes, $X^y$ is a set of test data (its size is 100) in the class $y$, $sim(x, y) = \frac{\sum_{l\in f}J_{p^l_x,p^l_y}}{|f|}$ represents the similarity between CDPs of $x$ and $y$ (see Equation~\ref{sec:jaccard}). Intuitively, the similarity between CDPs of the same class should be greater than the similarity between CDPs of two different classes.
}

Table~\ref{tab:distance} shows the average similarity between two CDPs in the same class (\ie, Column \textit{Intra}) and in different classes (\ie, Column \textit{Inter}) on the training data. The results show that the average similarity increases with the increase of the width, which further demonstrates that, if the CDP is too wide, it is not accurate to show the decision difference on two inputs. If the width of the CDP is too small, then it may miss some critical neurons (see Table~\ref{tab:singleinc}).



\begin{tcolorbox}[size=title]
	{\textbf{Answer to RQ1:}} 
CDPs are critical for the prediction while NCDPs are not critical, which reveals that the CDPs are more relevant to the internal decision logic of the DNN.
\end{tcolorbox}





\subsection{RQ2: Effectiveness of Path Abstraction}
\paragraph{Setup.} Since the abstract path represents the decision logic of training samples in the corresponding cluster, we first investigate whether the abstract path is still critical for the prediction of inputs in the cluster. Similar to RQ1, we mask the abstract CDP/NCDP and see the average inconsistency rate of training inputs in the cluster. Then, we evaluate whether the path abstraction could identify similar decision logic by measuring the similarity with or without clustering. 
\revise{ Specifically, for each cluster that contains a set of training samples $X$, we can calculate its abstract CDP based on all CDPs of $X$. To evaluate the precision of the path abstraction, for every input $x\in X$, we masked the same neurons (\ie, the abstract CDP) and evaluate the average inconsistency rate. Note that, in RQ1, we mask different neurons for different inputs of $X$ based on their individual CDPs while we mask the same neurons (in the abstract CDP) for all inputs of $X$ in RQ2. 
}

\subsubsection{Impact on prediction}
For the path abstraction of each DNN, we investigate a total of 16 configurations with different parameters $(k, \beta)$, where $k$ is the number of the clusters in each class and $\beta$ is the threshold to select neurons with high weights (the first column in Table~\ref{tab:abstractp}). Note that, when $k=1$, the abstract CDP is extracted by merging CDPs of all inputs that are predicted as the same class.

The results in Table~\ref{tab:abstractp} show that, with the increase of the number of clusters, the width tends to be slightly larger in most cases under the same threshold $\beta$ (see Column \textit{Wid.}), indicating that the average weight of neurons is also larger (\ie, CDPs in each cluster share more common neurons).
\revise{We count the number of the change of Inc.C and found that 81.25\% (39/48) of the Inc.C change is increasing or unchanged when the number of clusters increases. More fine-grained clustering (\ie, more clusters) could merge the similar CDPs together, by which the abstract CDP is more \textit{stable} (larger average weights) and \textit{accurate} (higher inconsistency rate after masking CDPs).
The results demonstrate the usefulness of clustering and merging.}
%


\begin{table}[!t]
    \centering
    \scriptsize
\setlength\tabcolsep{4pt}
    \caption{The average width and inconsistency rate (\%) after masking neurons in the abstract CDP and NCDP}
%
\begin{tabular}{ccccccccccccccccccc}
\toprule
\multirow{2}{*}{($k$, $\beta$)} & \multicolumn{3}{c}{SADL-1} & \multicolumn{3}{c}{SADL-2} & \multicolumn{3}{c}{VGG16(CIFAR-10)} & \multicolumn{3}{c}{AlexNet} & \multicolumn{3}{c}{\color{black}VGG16(ImageNet)}\tabularnewline
\cline{2-13} \cline{3-13} \cline{4-13} \cline{5-13} \cline{6-13} \cline{7-13} \cline{8-13} \cline{9-13} \cline{10-13} \cline{11-16} \cline{12-16} \cline{13-16} 
 & Wid. & Inc.C & Inc.NC & Wid. & Inc.C & Inc.NC & Wid. & Inc.C & Inc.NC & Wid. & Inc.C & Inc.NC & \color{black}Wid. & \color{black}Inc.C & \color{black}Inc.NC\tabularnewline
\midrule 
\midrule 
(1, 0.6) & 16.9 & 93.6 & 2.3 & 13.9 & 99.9 & 2.4 & 15.5 & 99.8 & 4.3 & 17.3 & 99.4 & 4.9 & \color{black}41.5	& \color{black}99.8	& \color{black}1.7\tabularnewline
(1, 0.7) & 14.6 & 89.2 & 8.4 & 9.4 & 99.8 & 2.4 & 13.1 & 99.3 & 4.3 & 13.6 & 99.1 & 13.9 & \color{black}34.4 &\color{black} 99.8 & \color{black}1.7\tabularnewline
(1, 0.8) & 12.5 & 78.8 & 29.9 & 6.4 & 99.2 & 2.4 & 10.8 & 98.8 & 4.3 & 9.8 & 96.3 & 16.3& \color{black}26.4 & \color{black}99.8 & \color{black}1.7\tabularnewline
(1, 0.9) & 10.4 & 73.6 & 59.8 & 4.1 & 90.9 & 2.4 & 7.9 & 100 & 10.1 & 5.7 & 50.9 & 37.6 & \color{black}17.7 & \color{black}99.8 &\color{black}1.7\tabularnewline
\midrule 
\midrule 
(4, 0.6) & 16.7 & 94.5 & 2.4 & 14.6 & 99.9 & 2.0 & 15.8 & 99.9 & 4.3 & \textbf{18.5} & \textbf{99.5} & \textbf{4.4}&\color{black}54.5&\color{black}100&\color{black}2.9\tabularnewline
(4, 0.7) & 15.4 & 95 & 2.5 & 10.8 & 99.9 & 2.0 & 13.5 & 99.9 & 4.8 & 15.1 & 99 & 5.8&\color{black}\textbf{49}&\color{black}\textbf{100}&\color{black}\textbf{1.9}\tabularnewline
(4, 0.8) & \textbf{14.6} & \textbf{94.1} & \textbf{3.9} & 7.6 & 99.6 & 2.0 & 11.2 & 99.9 & 4.8 & 11.8 & 96.4 & 16.2& \color{black}40.4&	\color{black}100	&\color{black}2.67\tabularnewline
(4, 0.9) & 12.2 & 88.1 & 14.3 & 4.8 & 95.3 & 2.0 & 8.4 & 100 & 9.2 & 7.6 & 80 & 26.7&\color{black}34.3	&\color{black}100&\color{black}2.6\tabularnewline
\midrule 
\midrule 
(7, 0.6) & 16.9 & 96.3 & 2.6 & 14.9 & 99.9 & 1.5 & 15.9 & 99.8 & 2.8 & 18.4 & 99.4 & 5.6&\color{black}64.2&\color{black}100&\color{black}4.3\tabularnewline
(7, 0.7) & 15.9 & 96.1 & 3.2 & 11.2 & 99.8 & 1.5 & 13.7 & 99.9 & 3.5 & 15.2 & 98.9 & 6&\color{black}59.5&\color{black}100&\color{black}2.9\tabularnewline
(7, 0.8) & 14.6 & 92.5 & 2.9 & 7.9 & 99.7 & 1.5 & 11.4 & 99.9 & 4.4 & 12.1 & 97.6 & 9.7& \color{black}51.8& \color{black}100& \color{black}3.6\tabularnewline
(7, 0.9) & 12.5 & 88.7 & 9.5 & \textbf{5.4} & \textbf{98.2} & \textbf{1.6} & \textbf{8.7} & \textbf{100} & \textbf{4.9} & 8 & 90.5 & 19&\color{black}46.7&\color{black}100	&\color{black}3.1\tabularnewline
\midrule 
\midrule 
(10, 0.6) & 16.9 & 95.3 & 1.7 & 15.2 & 99.9 & 1.1 & 15.9 & 99.9 & 2.0 & 18.5 & 99.4 & 5.2&\color{black}73.2&\color{black}100&\color{black}4.3\tabularnewline
(10, 0.7) & 16.1 & 95.1 & 3.1 & 11.5 & 99.8 & 1.2 & 13.7 & 99.9 & 2.7 & 15.4 & 98.9 & 6.3&\color{black}70.5&\color{black}100&\color{black}4.3\tabularnewline
(10, 0.8) & 14.6 & 93.0 & 4.3 & 8.3 & 99.8 & 1.3 & 11.5 & 99.9 & 3.4 & 12.4 & 98.4 & 8.5&\color{black}58.3&\color{black}100&\color{black}3.9\tabularnewline
(10, 0.9) & 13.5 & 89.2 & 8.8 & 5.4 & 98.8 & 1.3 & 8.7 & 99.9 & 8.8 & 8.6 & 92.9 & 17.7&\color{black}54.3&\color{black}100&\color{black}2.7\tabularnewline
\bottomrule 
\end{tabular}



    \label{tab:abstractp}
\end{table}

Compared with the results on single CDPs (Table~\ref{tab:singleinc}), we found that the abstract path could still achieve competitive performance and even better than single CDPs on some models. 
 It means that the path abstraction can be used to select commonly important neurons for all training data in a specific cluster (\ie, high inconsistency rate after masking CDPs and low inconsistency rate after masking NCDPs).
For example, on SADL-2, the inconsistency rates of CDP and NCDP are 98.2\% and 1.6\% ($k=7$), which are close to results of single CDPs (\ie, 100\% and 0\%) and the width is much smaller (\eg, 5.4\% for abstract CDPs and 27.2\% for single CDPs). For AlexNet, using the abstraction in the cluster, we could select more accurate CDPs (\eg, the average width of the abstract CDPs and the single CDPs is 18.5\% and 23.0\%).
Finally, in the following experiments, we select the configuration (4, 0.8), (7, 0.9), (7, 0.9), (4, 0.6) and (4, 0.7) as the parameters for SADL-1, SADL-2, VGG-16, AlexNet and VGG-16 (ImangeNet), respectively (the bold numbers). \qing{ These configurations are selected due to that we expect to select the neurons that can be representative and critical. Firstly, we could not leave critical neurons out, thus we need to ensure a great inconsistent rate under the selected configurations. Secondly, the width of critical neurons should not be great because the critical neurons are expected to be representative and discriminative. If the width is too large, there will be a large overlap between different critical neurons.}

\subsubsection{Decision Revealing}
Table~\ref{tab:clustersim} shows the results on the average similarity with and without clustering. Column \textit{\#Clus.} shows the number of clusters configured for each model. Without clustering, Column \textit{\#Intra\_Cla.} shows the average similarity between CDPs that are in the same class (\ie, the intra-class similarity). Column \textit{\#Inter\_Cla.} shows the average similarity between CDPs that belong to different classes (\ie, the inter-class similarity). After the clustering, we show the average similarity between CDPs in the same cluster (\ie, intra-cluster similarity shown in Column \textit{\#Intra\_Clus.}) and the average similarity between CDPs in different clusters of the \emph{same} class (\ie, inter-cluster similarity shown in Column \textit{\#Inter\_Clus.}). Note that inter-cluster similarity only compares clusters that belong to the same class

Comparing the results between intra-class similarity (without clustering) and intra-cluster similarity (with clustering), \qing{we found that the average similarity increases (\eg, from 0.628 to 0.708 in SADL-1)}. In particular, in the same class, the inter-cluster similarity is much smaller than intra-cluster similarity although the two clusters belong to the same class (\eg, 0.708 for intra-cluster and 0.381 for inter-cluster).
Moreover, the inter-cluster similarity is even close to the inter-class similarity. The results show that samples in the same class could have different decision logic (\eg, the black dogs and white dogs may be different). By the path abstraction, we generate fine-grained CDPs which could distinguish such differences well.

\begin{figure}
    \centering
     \includegraphics[width=0.90\linewidth]{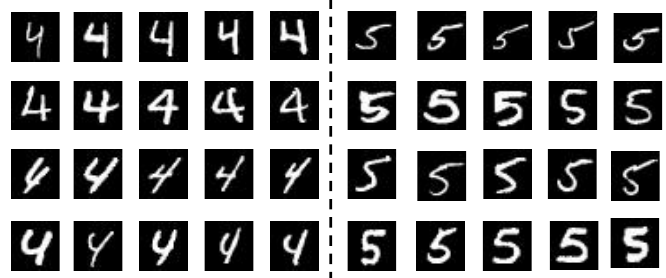}
    \caption{Inputs sampled from clusters in the class 4 and 5}
    \label{fig:clustervis}
\end{figure}

Fig.~\ref{fig:clustervis} shows some examples whose CDPs are clustered into different groups (on SADL-1). Each row represents one cluster.
We can observe that, from the human's perception, the inputs in the same cluster look very similar. For example, the digit ``4'' looks similar to the digit ``9'' in the fourth cluster and the digit ``5'' has a slight deflection in the first cluster. The results further show that inputs that trigger the similar decision logic is likely to be grouped into one cluster based on the extracted CDPs.

\begin{table}[!t]
    \centering
    \small
    \caption{The average similarity with and without clustering}
\begin{tabular}{cccccc}
\toprule 
Model & \#Clus. & Intra\_Cla. & Inter\_Cla. & Intra\_Clus. & Inter\_Clus.\tabularnewline
\midrule 
\midrule 
SADL-1 & 4 & 0.628 & 0.391 & 0.708 & 0.381\tabularnewline
SADL-2 & 7 & 0.486 & 0.232 & 0.524 & 0.260\tabularnewline
VGG16(CIFAR-10) & 7 & 0.671 & 0.320 & 0.703 & 0.316\tabularnewline
AlexNet & 4 & 0.478 & 0.300 & 0.524& 0.298 \tabularnewline
\color{black}VGG16(ImageNet) &\color{black}4	&\color{black}0.549&	\color{black}0.322&	\color{black}0.546&	\color{black}0.331\tabularnewline
\bottomrule 
\end{tabular}
    \label{tab:clustersim}
\end{table}

\begin{tcolorbox}[size=title]
	{\textbf{Answer to RQ2:}} 
\revise{Based on the aggregation (\ie, union) of all CDPs in each cluster}, the abstract CDP is still critical for predicting inputs in a cluster. Moreover, the abstract CDP is sensitive to reveal different decision logic. Inputs in the same cluster have similar decision and inputs in different clusters of the same class tend to have different decision behavior.
\end{tcolorbox}
\vspace{-.1in}
\subsection{RQ3: Sensitivity with Defect Detection}
\paragraph{Setup.} 
Motivated by~\cite{ncmislead,nerualmisled}, \qing{this experiment evaluates the effectiveness in detecting defects including natural errors\footnote{\qing{In this paper, "natural error" represents the natural input that is not manually crafted and is misclassified by the target model.}} and adversarial examples.} We follow the similar design in~\cite{nerualmisled}: first, we randomly select 1,000 benign samples (denoted as $s_0$) from the test data. 
Then, we construct several test suites by replacing a small number of samples of $s_0$ with the same number of errors such that they have the same size. 
To be specific, for natural errors, \qing{we select a certain number of inputs from the test data in each class}, which are predicted incorrectly. For adversarial examples, we adopt the PGD attack~\cite{madry2018towards} to generate a large number of adversarial examples, some of which are then randomly selected in each class. For natural errors and adversarial examples, we generate 6 test suites including different numbers of errors (\ie, 1\%, 2\%, 3\%, 5\%, 7\%, 10\%), respectively. We use $S$ to denote the set of the test suites including the initial set $s_0$.
\revise{To reduce the randomness, we repeat the process 5 times and calculate the average results.}

\begin{figure}
    \centering
     \includegraphics[width=1.1\linewidth]{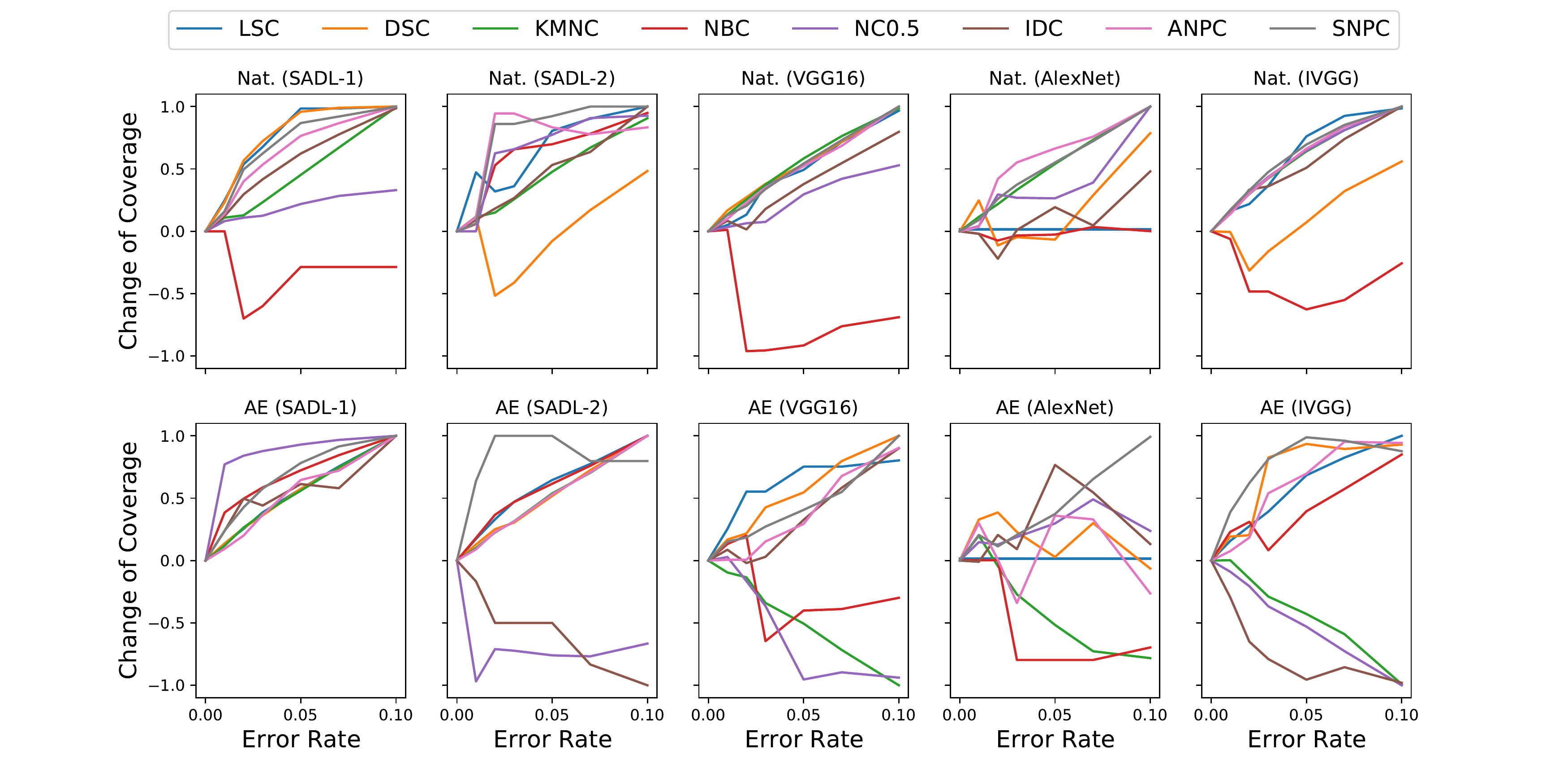}
    \caption{Coverage change on test \qing{suites} including different number of errors.}
    \label{fig:coveragechange}
\end{figure}

\paragraph{Results.} Fig.~\ref{fig:coveragechange} shows the results on the test suites including a different number of errors. 
To conduct the comparison between different approaches, all the coverage is normalized to [-1,1]. \revise{Specifically, for each test suite $s\in S$, we calculate its coverage change compared with the initial test suite $s_0$ with regard to the coverage criterion $Cov$, \ie,  $\Delta = \{Cov(s)-Cov(s_0)|  s\in S \}$.
For each $s\in S$, we normalize the change of the coverage as follows:

\begin{equation}
\label{equa:mutation}
NCov(s) = \frac{Cov(s)-Cov(s_0)}{max(\Delta)-min(\Delta)}
\end{equation}
\qing{Note that the coverage change can be a negative value}.

}

The first row shows the average results on natural errors (\emph{Nat.}) and the second row shows the results on adversarial examples (\emph{AE}). The horizontal axis represents the percentage of errors while the vertical axis represents the normalized coverage.

The results show that the coverage criteria exhibit different performance on different number of errors. \revise{In general, we could find that SNPC and ANPC tend to be more stable than other methods in terms of the sensitivity to errors.  In most cases, the trends of SNPC and ANPC are in line with the expectations, \ie, the coverage can increase when more errors (adversarial examples or natural errors) are introduced. NBC is the most unstable one because it mainly focuses on the boundary behaviors of the model. For KMNC and NC (with threshold 0.5), we observe that they perform better in natural errors than adversarial examples. The coverage of NC and KMNC increases when the number of natural errors increase. However, their coverage does not always increase on adversarial examples. Particularly, KMNC decreases on AEs of CIFAR-VGG16, ALexNet and ImageNet-VGG16. \qing{We conjecture that KMNC mainly considers the major functionality of the DNN while ignoring the boundary functionality}, thus replacing some benign samples with the corner cases (\eg, adversarial examples) could decrease the diversity of major behavior. LSC performs better in most cases but is insensitive in AlexNet, \ie, the coverage does not change on both adversarial examples and natural error of AlexNet, which indicates that 
the DNN may have a large impact on LSC. DSC is not sensitive on natural errors while it is better on adversarial examples. \qing{It is worth pointing out that most coverage criteria tend to have irregular changes in AE (AlexNet) while SNPC still has an increasing trend.}

Considering IDC that is also based on LRP, we can observe that the coverage increase is not as stable as SNPC and ANPC, \eg, on Nat.(AlexNet), AE (SADL-2), AE (AlexNet) and AE (IVGG). 
\qing{Although both IDC and NPC adopt LRP to identify the critical neurons layer by layer, NPC is more fine-grained than IDC, \ie, NPC performs the CDP-based clustering in each class. Thus, our results tend to be more stable.}

\qing{Moreover, we also observe that there are some correlations between these coverage. Specifically, the trend of coverage increase with regards to different coverage criteria could be similar.} For example, in Nat.(SDAL-1), LSC, DSC, ANPC and SNPC have the similar trend. In Nat.(SADL-2), SNPC and NC have the similar trend. Most of the coverage criteria have the similar trend in Nat.(VGG16), Nat.(IVGG) and AE(SADL-1). 
}

\begin{table*}[]
    \centering
    \caption{Correlation between coverage criteria and output impartiality}
\scriptsize
\begin{tabular}{ccccccccccccc}
\toprule 
 &  & KMNC & NBC & NC(0.0) & NC(0.2) & NC(0.5) & NC(0.75) & LSA & DSA &\color{black} IDC& ANPC & SNPC\tabularnewline
\midrule 
\midrule 
\multirow{5}{*}{size=100} & SADL-1 & -0.262 & -0.262 & -0.300 & 0.099 & 0.315 & 0.001 & 0.000 & -0.638 & \color{black}-0.360& {-0.244} & \textbf{0.723}\tabularnewline
 & SADL-2 & -0.029 & -0.029 & \textbf{0.553} & -0.397 & -0.065 & -0.111 & 0.000 & 0.181 &\color{black} \textbf{0.587} & \textbf{0.482} & \textbf{0.789}\tabularnewline
 & VGG16 & -0.348 & -0.348 & 0.037 & -0.031 & -0.210 & -0.273 & 0.000 & -0.145 & \color{black} 0.052& 0.012 & -0.055\tabularnewline
 & AlexNet & -0.457 & \textbf{0.593} & 0.000 & 0.000 & 0.000 & 0.000 & 0.000 & -0.320 & \color{black} \textbf{0.912} & \textbf{0.961} & \textbf{0.989}\tabularnewline
 \cline{2-13}
 & Avg. & -0.274 & -0.011 & 0.072 & -0.082 & 0.010 & -0.096 & 0.000 & -0.230 &\color{black} \textbf{0.298 } & \textbf{0.601} & \textbf{0.612}\tabularnewline
\midrule 
\midrule 
\multirow{5}{*}{size=500} & SADL-1 & \textbf{0.639} & -0.116 & 0.000 & 0.000 & 0.000 & 0.000 & 0.000 & -0.639 & \color{black} -0.707& \textbf{0.340} & \textbf{0.870}\tabularnewline
 & SADL-2 & \textbf{0.543} & 0.041 & 0.000 & 0.000 & 0.000 & 0.000 & 0.000 & 0.330 &\color{black} \textbf{0.473} & \textbf{0.584} & \textbf{0.817}\tabularnewline
 & VGG16 & \textbf{0.414} & \textbf{0.589} & 0.000 & 0.000 & 0.000 & 0.000 & 0.000 & -0.255 &\color{black} \textbf{ 0.210}& \textbf{0.141} & \textbf{0.821}\tabularnewline
 & AlexNet & \textbf{0.539} & \textbf{0.586} & 0.000 & 0.000 & 0.000 & 0.000 & 0.000 & -0.170 & \color{black}\textbf{0.931}&\textbf{0.908} & \textbf{0.974}\tabularnewline
  \cline{2-13}
 & Avg. & \textbf{0.534} & 0.275 & 0.000 & 0.000 & 0.000 & 0.000 & 0.000 & -0.184 & \color{black} \textbf{0.227} & \textbf{0.893} & \textbf{0.871}\tabularnewline
\bottomrule 
\end{tabular}
    \label{tab:corr}
\end{table*}



\begin{tcolorbox}[size=title]
	{\textbf{Answer to RQ3:}} 
    Overall, ANPC and SNPC are more sensitive to different errors than other coverage criteria due to the CDP awareness. Different coverage criteria could achieve similar increasing trend, indicating that they tend to have some correlation.
    
\end{tcolorbox}
\vspace{-.1in}
\subsection{RQ4: Correlation with Output Impartiality}
\revise{
As described in~\cite{ncmislead,test:deephunter,alshahwan2014coverage}, the test data or errors should be diverse in terms of the output.
The existing work~\cite{sekhon2019towards} has demonstrated that the coverage (\eg, neuron coverage) can be improved by a few samples. However, a test suite that only contains one category of samples (\eg, dog) is not enough to comprehensively reveal diverse defects of the DNN (\eg, the errors in other categories). In addition, the coverage criteria are usually used to guide the test case generation. The coverage criteria that are sensitive to output diversity can guide the testing tools to generate more diverse test cases belonging to different categories while the insensitive coverage criteria may generate biased test cases~\cite{test:deephunter}. Hence, we evaluate the correlation between the coverage criteria and the output diversity in this section.
}
\paragraph{Setup.} 
We follow the same setting in~\cite{ncmislead} and use the proposed tool to generate a number of test suites, which have the same size. 
\revise{Specifically, it mainly contains three steps:
\begin{enumerate}[leftmargin=*]
    \item For each dataset, we randomly select two seed test suites with different size (\ie, 100 and 500).
    \item Based on each seed test suite, we adopt the tool~\cite{ncmislead} to generate new test suites that have the same size with the seed test suite (\ie, generate new input by perturbing each input of the seed test suites). The tool~\cite{ncmislead} extends the existing adversarial attacks (\eg, C\&W ~\cite{cw2017} and PGD~\cite{madry2018towards}) with the divergence function such that more diverse test cases can be generated. By setting different configurations (\eg, diversity regularization weight, confidence), multiple test suites 
   can be generated. For each seed test suite on SADL-1, SADL-2, VGG-16 (CIFAR-10) and AlexNet, we generate 48, 171, 167 and 168 new test suites\footnote{The number of test suites is different because it depends on the number of configurations in~\cite{ncmislead} (\ie, a new test suite is generated by a configuration), where one variable in the configurations is the layers of the neuron network.}, respectively. 
   Note that, we did not include the results of ImageNet in this experiment because all adversarial examples generated by the tool~\cite{ncmislead} belong to the same class, which makes all correlation scores as NaN.
    \item For a generated test suite, we calculate the coverage and the output diversity, respectively. Finally, we calculate the \textit{Pearson} correlation between them.
\end{enumerate}
} 

Specifically, we use the output impartiality metric~\cite{ncmislead} for a test suite $T$:
$$
output\_impartiality(T) = \frac{\sum_{t\in C_k}P_{t=C_k} logP_{t=C_k} }{\frac{1}{|c|}log\frac{1}{|c|}}
$$
where $|C|$ is the cardinality of classes and $P_{t=C_k}$ represents the
percentage of the test cases $t$ predicted to belong to class $C_k$. 


\paragraph{Results.} Table~\ref{tab:corr} shows the correlation on different models. The overall results demonstrated that, compared with existing coverage criteria, ANPC and SNPC are more strongly correlated with the output impartiality. In particular, we have the following findings:
1) Compared with the test suite with smaller size (\eg, 100),  the correlation can be stronger when the size of the test cases becomes larger (\ie, 500). \qing{For example, for the size 500, ANPC and SNPC are positively and strongly correlated with the output impartiality in all models while there are some negative correlations in the size 100}. KMNC is also positively correlated with output impartially when the size is 500. 2) \qing{Considering the results of ANPC and SNPC, we found that SNPC is more strongly correlated with output comparability.} It is because ANPC is more fine-grained and could capture the minor activation difference (\ie, coverage increase) even when they have output bias (\ie, the same prediction output).

\qing{For the results that ANPC and SNPC achieve negative correlation (\eg, -0.244 on SADL-1, -0.055 on VGG16) when the size of the test suite is 100,  our in-depth analysis reveals that, for some test suites with smaller size,  the prediction outputs of the test cases are biased (\eg., these test cases are predicted as the same class), which caused the low output impartiality. However, these samples can still trigger different CDPs (\ie, different similarities compared with the corresponding CDP), which lead to high coverage. Hence, the correlation is negative. When the size becomes larger (\eg, 500), the test suite becomes more stable (\ie, the prediction results are less biased), which makes that the coverage is positively correlated with the output impartiality.
 }

\revise{We also observe that the correlation of IDC is higher than other baselines but it is still less than ANPC and SNPC. The average correlation scores of IDC are less than ANPC and SNPC. For example, when the size is 100, the average score of IDC is 0.298 while the average scores of ANPC and SNPC are 0.601 and 0.612, respectively. Except for VGG16 (size=100), all correlation scores of SNPC are larger than IDC. The results indicate the effectiveness of our fine-grained coverage criteria.}

\begin{tcolorbox}[size=title]
	{\textbf{Answer to RQ4:}}
Generally, ANPC and SNPC are positively correlated with the output impartiality.
The model and the size of test suite could affect the correlation.
\end{tcolorbox}
\vspace{-.1in}

\subsection{RQ5: Efficiency of NPC}
We use the same dataset from RQ3 and compare the time overhead of calculating the coverage of a test suite (\ie, 1000 samples). Note that, except for NC, all other coverage criteria depend on the training samples and there are some prepossessing process. For example, KMNC and NBC need to profile the training data. LSC and DSC need to calculate all activation values of the training data. ANPC and SNPC need to calculate the CDPs for all training data. IDC also requires the LRP calculation for all training data. Since the prepossessing phase can be executed only once and it is usually calculated offline, we ignore the time of the prepossessing and only compare the coverage calculation time for a given test set. 

Table~\ref{tab:time} shows the results. We found that, NC (with the threshold 0.0) is the most efficient one as it only depends on the neuron output. For ANPC and SNPC, we list the total time and the CDP path extraction time (the number in brackets).The results show that the path extraction is efficient. Compared with others, ANPC and SNPC take more time since more information are required (\ie, the CDP calculation). IDC is more efficient because it only calculates the distance between the 
critical neuron values of the test cases and the training samples at one layer.
Actually, it is a trade-off between effectiveness (\ie, leveraging more information) and the efficiency. 

\begin{tcolorbox}[size=title]
	{\textbf{Answer to RQ5:}}
Although ANPC and SNPC adopt the interpretation analysis, the time overhead is not expensive. The CDP extraction is efficient and it only takes a proportion in the total time overhead of the coverage calculation. 
\end{tcolorbox}
\vspace{-.1in}

\begin{table}[!t]
    \centering
    \small
    \caption{The time overhead of coverage criteria on different models (seconds)}

\begin{tabular}{ccccccccc}
\toprule 
 & ANPC & SNPC & KMNC & NBC & LSC & DSC & NC(0.0) & IDC\tabularnewline
\midrule 
\midrule
SADL-1 & 10.25(2.60) & 8.83(2.60) & 3.50 & 1.92 & 26.25 & 59.75 & 0.67 & 1.11\tabularnewline
SADL-2 & 39.50(14.45) & 47.33(14.45) & 26.75 & 13.67 & 24.67 & 166.17 & 1.33& 0.82\tabularnewline
VGG16(CIFAR-10) & 135.50(35.10) & 91.17(35.10) & 54.67 & 28.75 & 21.00 & 73.00 & 2.17& 2.81\tabularnewline
AlexNet & 57.00(13.35) & 29.50(13.35) & 104.33 & 53.25 & 37.08 & 136.00 & 1.08& 1.21\tabularnewline
\color{black}VGG16(ImageNet) & \color{black}203.24(59.67) & \color{black}119.75(59.67) & \color{black}146.43 & \color{black}88.02 & \color{black}154.27 & \color{black}173.27 & \color{black}30.45 &  \color{black} 13.21\tabularnewline
\bottomrule 
\end{tabular}

    \label{tab:time}
\end{table}

\section{Threats to Validity and Discussion}\label{sec:limit}
\paragraph{Threats to Validity}
\revise{
The internal threats to validity lies in our implementations including the LRP implementation, the path extraction, the path abstraction, the coverage criteria and the implementation in each experiment. To reduce this threat, the co-authors carefully checked the correctness of our implementation. 

The external threats to validity include the selection of the subjects and the tools used in the evaluation. Specifically, the selection of datasets and the used models could be a threat. To reduce this threat, we selected four widely-used subjects (MNIST, CIFAR-10, SVHN and ImageNet) and 4 DNNs (SADL-1, SADL-2, VGG-16 and AlexNet), where ImageNet is a large-scale dataset. Another threat could be the generated data including adversarial examples, the selected natural errors and test suites (in RQ3, RQ4 and RQ5). To mitigate this threat, we follow the same setting in the existing work~\cite{ncmislead,nerualmisled} and generate diverse data with multiple configurations. In addition, the usage of the existing tool~\cite{ncmislead} can be a threat. Since our method, the used models and the tool~\cite{ncmislead} are developed by PyTorch, we manually re-implement the tool~\cite{test:surprise,IDCcov} by PyTorch, which may not the exactly the same with the original version.  We integrate our method, LSC, DSC, NC, KMNC and NBC into the tool~\cite{ncmislead} to evaluate the correlation.
We carefully checked the code to mitigate the threat.

The construct threats mainly lie in the randomness, the selected baselines, the parameters. First, one threat is the randomness in the selection of the data. To mitigate the treat, in RQ3, we repeat each experiment 5 times (\eg, the generation of adversarial examples) and calculate the average results. In RQ4, we follow the setting~\cite{ncmislead} and generate a large number of test suites to reduce the randomness. Second, we adopted the state-of-the-art approaches that have been demonstrated to be insufficient in DL testing. We compare our method with them to demonstrate the advantages of our proposed coverage criteria. Thirdly, the hyper-parameters in our methods and the baselines could be a threat to validity.
To reduce the threat, for the baselines, we follow the settings in the existing works~\cite{ncmislead,test:surprise,ma2018deepgauge}. For our method, we set multiple configurations in terms of the hyper-parameters $\alpha$, $\beta$ and the number of clusters $k$. However, in RQ3 and RQ4, we select the same values of $\alpha$, $\beta$ and $k$ for all paths and classes. Actually, we can select the optimal configurations by adopting different $\alpha$, $\beta$ and $k$. In the future, we plan to evaluate our method with more configurations.
}



\paragraph{Limitation}
Like the coverage criteria in traditional software, we think there is no one best coverage criterion and each one has some limitations. The main advantage of NPC lies in its interpretability and the clear relationship with the internal decision of the DNN.
\qing{In this section, we summarize our limitations and discuss the potential solutions in the future.}
\begin{itemize}
    \item Currently, our approach mainly considers the classification since the current interpretation technique (\ie, LRP) does not support the DNN with regression prediction. \qing{Even though the used interpretation technique is orthogonal to our approach, our approach is general as long as the critical neurons could be calculated.}  We believe the interpretation techniques that work on regression tasks in the future will further solve this limitation.
    \item Our approach mainly works on feed-forward neural networks. For recurrent neural networks (RNNs), whose number of iterations is not fixed, it could be challenging for the interpretation technique (\eg, relevance calculation in LRP). In the future, we may extend LRP by unrolling the RNN to address this limitation.
    \item \revise{There is another limitation on the selection for the number of the clusters during the path abstraction. Currently, we select the same number for all classes, which may not be the optimal setting. In the future, we will investigate the automated selection on this parameter.}
\end{itemize}

\section{Related Work}\label{sec:realted}
This section summarizes the related work on DL testing and adversarial attacks.
\subsection{DL Testing}
DeepXplore \cite{pei2017deepxplore} proposes the first white-box coverage criteria, \ie, Neuron Coverage, which calculates the percentage of activated neurons. A differential testing approach is proposed to detect the errors by increasing NC. \qing{DeepGauge \cite{ma2018deepgauge} then extends NC and proposes a set of more fine-grained coverage criteria by considering the distribution of neuron outputs from training data. }

Inspired by the coverage criteria in traditional software testing, some coverage metrics~\cite{sun2018concolic,ma2019deepct,ma2018deepmutation} are proposed. DeepCover \cite{sun2018concolic} proposes the MC/DC coverage of DNNs based on the dependence between neurons in adjacent layers. 
 DeepCT~\cite{ma2019deepct} adopts the combinatorial testing idea and proposes a coverage metric that considers the combination of different neurons at each layer. 
DeepMutation \cite{ma2018deepmutation} adopts the mutation testing into DL testing and proposes a set of operators to generate mutants of the DNN.
Furthermore, \qing{Sekhon \etal \cite{sekhon2019towards} analyzed the limitation of exiting coverage criteria and proposed a more fine-grained coverage metric that considers both of the relationships between two adjacent layers and the combinations of values of neurons at each layer.} 

Based on the neuron coverage, DeepPath~\cite{wang2019deeppath} initially proposes the path-driven coverage criteria, which considers the sequentially linked connections of the DNN. Although both of DeepPath and our method focus on path-based coverage, the approach and the underlying principles are totally different. The path in DeepPath is the \textit{syntactic} connections between neurons in different layers and the semantics of the path is unknown. Differently, we adopt the interpretation techniques to define paths, in which the neurons are selected based on the relevance with the decision. Thus, CDPs have more clear \textit{semantics}, \ie, they are related to the decision logic. 

The aforementioned techniques mainly consider syntactic structure of DNN, which is black-box and hard to understand. Differently, Kim \etal \cite{test:surprise} proposed the coverage criteria that measure the surprise of the inputs. The assumption is that surprising inputs introduce more diverse data such that more behaviors could be tested. Surprise metric measures the surprise score by considering \textit{all} neuron outputs of one or several layers. It is still unclear how the surprise coverage (calculated from some layers) is related to the decision logic. Differently, our method not only selects the \textit{critical} neurons at each layer but also considers the relationships between layers.

Based on the these criteria, some automated testing techniques \cite{pei2017deepxplore,tian2018deeptest,test:deephunter,odena2018tensorfuzz,sun2018concolic,test:deepstellar,zhang2020towards} are proposed to generate test inputs towards increasing the coverage.
In addition, while the coverage criteria are widely studied, the existing work~\cite{dong2019limited,ncmislead,nerualmisled,sekhon2019towards,fsecorrelation,DeepGini} found some preliminary evidence about the limitation of the \textit{structural} criteria. For example, the coverage is too coarse and is not correlated to the defect detection and robustness. Such findings motivate this work that proposes \textit{interpretable} coverage criteria by extracting the semantic structure (\ie, the decision graph). 

Recently, some coverage criteria~\cite{decisiontreeCov,IDCcov} are also proposed based on the decision of the DNN. The technique~\cite{decisiontreeCov} adopts the explanation technique LIME~\cite{lime} to extract a decision tree that explains the decision of the DNN on an input. Based on the decision tree, the symbolic execution is used to generate test cases that can maximize the path coverage on the decision tree. Different with~\cite{decisiontreeCov}, we extract the decision graph that could represent the global decision behaviors of the DNN. \qing{Moreover, in~\cite{decisiontreeCov}, it mainly focuses on the problem of fairness evaluation while it is hard to extract such a decision tree on the high dimensional data such as image domain.} IDC~\cite{IDCcov} adopts the interpretation technique to select the important neurons in one layer. Based on the training data, it then groups the activation values of important neurons into a set of clusters and uses the clusters to measure the coverage. Similar with our technique, IDC also selects the critical neurons. Differently, NPC considers the relations between layers and proposes the path-based coverage instead of only one layer. \qing{Thus, NPC does not need to select one of the target layers. In addition, we propose two different path coverage criteria from the control flow and data flow of the DNN. Significantly, there are also two papers \cite{senn} and \cite{litl} extracting critical neurons to represent decision behaviors. However, they focus on adversarial attacks and 
there are relatively larger computation costs. } Except for the testing on the trained model, some techniques (\eg,~\cite{xiao2021self}) have been proposed to build the self-checking system that can monitor DNN output and trigger an alarm if the output is likely to be incorrect after the model is deployed.
\qing{For more relevant discussions on the recent progress of machine learning testing, we refer the interesting readers to the recent comprehensive survey~\cite{zhang2020machine}.}

\subsection{Adversarial Attacks}
Recently, there are many adversarial attack techniques such as adversarial noise attacks including \emph{FGSM}~\cite{Goodfellow2015}, \emph{JSMA}~\cite{papernot2016SP}, \emph{BIM}~\cite{Kurakin2017adver}, \emph{DeepFool}~\cite{Seyed2016DeepFool},  and \emph{C\&W}~\cite{cw2017}, and natural degradation-based adversarial attacks including \emph{ABBA}~\cite{guo2020watch}, \emph{SPARK}~\cite{guo2020spark}, 
\emph{AVA}~\cite{tian2021ava}, 
and \emph{Pasadena}~\cite{cheng2021pasadena}.
The adversarial examples are generated by adding visually imperceptible perturbation to an input such that the new input is misclassified. It is worth to think and discuss about the difference between deep learning testing and adversarial attacks. \qing{The common thing is that both the testing and attack could test the robustness of the model by generating incorrect inputs.} Differently, the adversarial attack is ad-hoc and generates the \textit{specific} adversarial examples based on gradient calculation. 
DL testing aims to test the DNN more systematically by exploring more decision logic of the model. Thus, the errors generated by DL testing are expected to be more diverse (\ie, triggering different logic) than adversarial attacks. For example, given one input, the attack techniques usually generate the similar adversarial examples while DL testing could generate different adversarial examples with the guidance of the coverage criteria.

\section{Conclusion}\label{sec:con}
Motivated by the recent findings \cite{ncmislead,nerualmisled,dong2019limited} about the limitation of the existing DL testing criteria, this paper proposes the neuron path coverage via extracting the decision structure of the targeted DNN. For a DNN, we extract the decision logic based on the critical decision paths from the training data. \qing{Two path-based coverage criteria are then proposed to measure whether new decision logic is covered by the test cases. Specifically, SNPC is designed based on the control-flow of the DNN which measures whether different critical neurons are covered; ANPC is designed based on the data-flow of the DNN which measures the activation values of the critical neurons.} To the best of our knowledge, this is the first work that proposes the interpretable coverage criteria from the control-flow and data-flow of the DNN. The evaluation results demonstrated the effectiveness of the CDP in the decision revealing of the DNN and the usefulness of the coverage criteria.

\section*{Acknowledgments}
%
This research is partially supported by the National Research Foundation, Singapore under its the AI Singapore Programme (AISG2-RP-2020-019), the National Research Foundation, Prime Ministers Office, Singapore under its National Cybersecurity R\&D Program (Award No. NRF2018NCR-NCR005-0001), NRF Investigatorship NRFI06-2020-0022-0001,  the National Research Foundation through its National Satellite of Excellence in Trustworthy Software Systems (NSOE-TSS) project under the National Cybersecurity R\&D (NCR) Grant Award No. NRF2018NCR-NSOE003-0001, the Ministry of Education, Singapore under its Academic Research Fund Tier 3 (MOET32020-0004), the JSPS KAKENHI Grant No.JP20H04168, JP19K24348, JP19H04086, JP21H04877 and JST-Mirai Program Grant No.JPMJMI20B8, Japan. Lei Ma is also supported by Canada CIFAR AI Program and Natural Sciences and Engineering Research Council of Canada. Any opinions, findings and conclusions or recommendations expressed in this material are those of the author(s) and do not reflect the views of the Ministry of Education, Singapore.








\small
\bibliographystyle{ACM-Reference-Format}
  \bibliography{ref}
\end{document}